\documentclass{article}

\usepackage[preprint]{corl_2026} 

\usepackage[utf8]{inputenc} 
\usepackage[T1]{fontenc}    
\usepackage{hyperref}       
\usepackage{url}            
\usepackage{booktabs}       
\usepackage{amsfonts}       
\usepackage{nicefrac}       
\usepackage{microtype}      
\usepackage[table]{xcolor}         
\usepackage{graphicx}
\usepackage{wrapfig, blindtext}

\usepackage{algorithmic}
\usepackage{algorithm}

\usepackage{amsmath}
\usepackage{amssymb}
\usepackage{mathtools}
\usepackage{amsthm}

\usepackage{bm}
\usepackage{bbm}
\usepackage{multirow}
\usepackage{subcaption}
\usepackage{pifont}
\usepackage{makecell}
\usepackage{comment}
\usepackage{enumitem}

\newtheorem{theorem}{Theorem}[section]
\newtheorem{proposition}[theorem]{Proposition}

\title{Physics-informed Goal-Conditioned Reinforcement Learning under Hybrid Contact Dynamics}

\author{%
  Vittorio Giammarino, Anastasios Manganaris, and Ahmed H. Qureshi \\
  Department of Computer Science\\
  Purdue University\\
  \texttt{\{vgiammar,amangana,ahqureshi\}@purdue.edu} \\
}

\begin{document}
\maketitle


\begin{abstract}
Learning to reach arbitrary goals from sparse feedback requires agents to infer a rich notion of reachability across state--goal pairs. Goal-conditioned reinforcement learning (GCRL) tackles this challenge by learning policies that generalize across goals, but this generalization becomes increasingly difficult as the underlying dynamics become high-dimensional, hybrid, or contact-dependent. To address this issue, physics-informed GCRL (Pi-GCRL) introduces optimal-control-inspired inductive biases into goal-conditioned value learning. While Pi-GCRL methods have proven effective in navigation and object-free goal-reaching domains, their reliability in contact-rich tasks remains unclear, where contact interactions induce hybrid dynamics, mode-dependent controllability, and nonsmooth value landscapes. In this work, we show that these structural properties can cause existing Pi-GCRL methods to degrade when applied naively to contact-rich manipulation. Motivated by this analysis, we introduce contact-aware and hierarchical formulations that apply physics-informed inductive biases selectively across the manipulation problem. Our results provide a principled step toward extending Pi-GCRL to contact-rich manipulation.
\end{abstract}


\vspace{-0.2cm}
\section{Introduction}
\vspace{-0.2cm}

Recent progress in robot learning has been driven largely by imitation-based methods, from behavior cloning~\cite{pomerleau1988alvinn} and diffusion policies~\cite{chi2025diffusion} to vision-language-action models trained on large-scale data~\cite{black2024pi_0, shukor2025smolvla}. While these methods benefit from strong supervision, their performance remains tied to the quality, diversity, and coverage of the collected demonstrations. As a result, we still require solutions that enable agents to improve and adapt beyond the demonstrated behavior.

Reinforcement Learning (RL) provides a natural framework for this goal, as it allows policies to improve through interaction with task rewards~\cite{chen2025pi}. These rewards supervise the learning process by defining desirable outcomes. However, they can be difficult to design in complex environments, often requiring substantial fine-tuning, simulation access, and trial-and-error~\cite{ibrahim2024comprehensive}. Sparse rewards offer a simpler alternative, but they can also exacerbate the complexity of the learning process~\cite{giammarino2024combining}. This tension has motivated increasing interest in Goal-Conditioned RL (GCRL)~\cite{kaelbling1993learning, schaul2015universal,andrychowicz2017hindsight}, which shifts the focus from designing task-specific rewards to learning from sparse goal-reaching signals.

A central challenge in GCRL is the weak supervision available for learning the geometry of the underlying control problem. Specifically, using only sparse, indirect feedback, GCRL algorithms must estimate which states can be connected through feasible transitions and how costly those transitions are. From a graph-search perspective, states can be viewed as nodes, feasible transitions as edges, and goal reaching amounts to estimating shortest-path distances over the graph induced by the system dynamics~\cite{yang2020plan2vec, wang2023optimal}.

Physics-informed GCRL (Pi-GCRL) directly addresses this reachability learning problem by regularizing value functions with partial differential equations (PDEs) inspired by optimal control~\cite{settai2026a, giammarino2026physicsinformed, giammarino2026goal}. Rather than learning an arbitrary function over state-goal pairs, Pi-GCRL encourages the value landscape to behave like a physically meaningful cost-to-go. In this sense, it provides a value-shaping inductive bias that preserves the sparse-reward formulation while aligning value learning with the underlying control structure.

So far, Pi-GCRL has shown promising results in navigation and object-free goal-reaching tasks. However, its effectiveness in contact-rich robotic manipulation remains unclear~\citep{giammarino2026physicsinformed, giammarino2026goal}. This gap is important because manipulation differs qualitatively from standard goal-reaching problems~\citep{manipulation}. In contact-rich settings, the dynamics are hybrid, as the system alternates between free-space motion, contact-mediated interaction, and object transport. These mode transitions can induce state-dependent changes in controllability, non-smooth value landscapes and sharp changes in local geometry, potentially invalidating the assumptions under which PDE constraints provide useful inductive biases. In this paper, we shed light on this failure mode and derive novel formulations that enable reliable Pi-GCRL in contact-rich manipulation. In summary, our contributions are:
\begin{itemize}[leftmargin=*]
    \item \textbf{Principled analysis of Pi-GCRL under hybrid contact dynamics.}
    We analyze why physics-informed value regularization can become misaligned with contact-rich manipulation. In particular, we show that mode-dependent controllability can make PDE constraints incompatible with the dynamics of the system; explaining why previously developed methods can degrade when applied directly to full state spaces.

    \item \textbf{Contact-aware and hierarchical formulations.}
    Guided by this analysis, we derive two design principles for applying physics-informed constraints in manipulation: $(i)$ constrain value gradients along feasible directions, and $(ii)$ apply PDE-based regularization only in fully controllable representations. We instantiate these principles through a contact-aware residual and a hierarchical decomposition that separates high-level object-centric reasoning from low-level goal reaching.

    \item \textbf{Empirical validation across offline goal-conditioned manipulation.}
    We validate our theoretical and algorithmic claims across contact-rich manipulation tasks, showing that the proposed formulations improve over naive full-state regularization. We further provide preliminary experiments with real-world robot data, demonstrating that the framework can be extended to real-world settings.
\end{itemize}

\vspace{-0.2cm}
\section{Related work}

\vspace{-0.2cm}
\paragraph{GCRL}
GCRL conditions policies on goals, enabling agents to reuse goal-reaching experience across tasks~\citep{kaelbling1993learning, schaul2015universal}. This perspective has led to a broad family of methods for learning from sparse feedback, ranging from hindsight relabeling~\citep{andrychowicz2017hindsight}, contrastive objectives~\citep{eysenbach2022contrastive}, and state-occupancy matching~\citep{ma2022offline}, to hierarchical approaches that decompose long-horizon problems into shorter subtasks~\citep{giammarino2026goal, park2024hiql, haramati2026hierarchical, park2026horizon, ahn2026option}. In the offline setting, where policies must be learned from fixed datasets, this challenge has motivated specialized algorithms for offline goal reaching~\citep{park2024hiql, chebotar2021actionable, yang2022rethinking, yang2023essential, mezghani2023learning, sikchi2023smore}. A complementary perspective interprets optimal goal-conditioned value functions as shortest feasible paths, inducing an asymmetric distance-like structure over the state space~\citep{wang2023optimal, sontag1995abstract, liu2023metric}. Quasimetric RL (QRL)~\citep{wang2023optimal} formalizes this idea by constraining value functions to quasimetric parameterizations aligned with the dynamics-induced geometry of goal reaching~\citep{liu2023metric, pitisinductive, durugkar2021adversarial}.

\vspace{-0.3cm}
\paragraph{Physics-informed RL}
Recent work has introduced physics-informed biases into value learning by regularizing value functions with PDE-based constraints inspired by continuous-time optimal control. A few studies propose Hamilton-Jacobi-Bellman (HJB)-style constraints for online or offline RL~\citep{settai2026a, lienenhancing}, demonstrating the potential of PDE-based value regularization. However, these approaches do not focus on GCRL and do not address the specific challenges of contact-rich robotic manipulation. Within GCRL, \citet{giammarino2026physicsinformed} propose Eikonal regularization for goal-conditioned value learning~\citep{noack2017acoustic}, using the Eikonal equation as a computationally convenient surrogate for the HJB equation. Building on this idea, \citet{giammarino2026goal} combine Eikonal regularization with quasimetric value parameterizations, yielding an Eikonal-constrained version of QRL~\citep{wang2023optimal}. While these methods show strong results in navigation and object-free goal-reaching, their direct application to contact-rich manipulation remains limited. Our work focuses on this gap. We formally characterize when the Eikonal equation provides a sound inductive bias for goal-conditioned value learning and why it can break down under hybrid dynamics and state-dependent controllability. This focus distinguishes our work from recent HJB-inspired alternatives for GCRL~\citep{viswanath2026physics}: we do not seek to replace Eikonal regularization in general, but instead analyze its limitations in contact-rich manipulation and develop formulations tailored for this setting.

\vspace{-0.3cm}
\paragraph{GCRL for robotic manipulation}
GCRL has also become a natural framework for robotic manipulation, particularly as language-conditioned policies can be interpreted as goal-conditioned policies with goals specified through language or language-aligned embeddings~\citep{stepputtis2020language, ma2022vip, ma2023liv}. Recent work has further integrated pretrained vision-language models into GCRL~\citep{wang2025versatile}, generated task-specific policy parameters from goal descriptions~\citep{zhou2026textithypergoalnet}, used automata for temporally extended manipulation~\citep{manganaris2026automaton}, and developed hierarchical entity-centric methods for long-horizon robotic tasks~\citep{haramati2026hierarchical}. In contrast, our work studies PDE-constrained goal-conditioned value learning and develops formulations tailored to contact-rich manipulation.

\vspace{-0.2cm}
\section{Preliminaries}

\vspace{-0.2cm}
\paragraph{GCRL formulation} 
We consider a finite-horizon discounted goal-conditioned Markov Decision Process (MDP) defined by $(\mathcal{S}, \mathcal{G}, \mathcal{A}, \mathcal{T}, \mathcal{R}, \mathcal{P}_g, \rho_0, \gamma)$, where $\mathcal{S}$ is the state space, $\mathcal{G}$ the goal space, $\mathcal{A}$ the action space, and $\mathcal{T}: \mathcal{S} \times \mathcal{A} \to \mathcal{P}(\mathcal{S})$ the transition map, with $\mathcal{P}(\mathcal{S})$ denoting the set of probability distributions over $\mathcal{S}$. The goal-conditioned reward function $\mathcal{R}: \mathcal{S} \times \mathcal{G} \to \mathbb{R}$ is defined through a goal-satisfaction set $\mathcal{S}_g \subseteq \mathcal{S}$ as $\mathcal{R}(s,g)=0$ if $s \in \mathcal{S}_g$ and $\mathcal{R}(s,g)=-1$ otherwise. At the beginning of each episode, a goal $g \sim \mathcal{P}_g$ is sampled from the goal distribution $\mathcal{P}_g \in \mathcal{P}(\mathcal{G})$, and an initial state $s_0 \sim \rho_0$ from the initial state distribution $\rho_0 \in \mathcal{P}(\mathcal{S})$. The agent seeks to maximize $J(\pi) = \mathbb{E}_{g \sim \mathcal{P}_g, \tau_\pi(g)} \left[ \sum_{t=0}^{T} \gamma^t \mathcal{R}(s_t,g) \right]$, where $\gamma \in (0,1]$ is the discount factor and $\tau_\pi(g) = (g,s_0,a_0,s_1,a_1,\dots,s_T)$ denotes a trajectory generated by the policy $\pi: \mathcal{S} \times \mathcal{G} \to \mathcal{P}(\mathcal{A})$. The value function induced by $\pi$ is defined as $V^\pi(s,g) = \mathbb{E}_{\tau_\pi(g)} \left[ \sum_{t=0}^{T} \gamma^t \mathcal{R}(s_t,g) \mid S_0=s,\, G=g \right]$. Finally, we denote parameterized functions by $\pi_{\bm{\theta}}$, with parameters $\bm{\theta} \in \varTheta \subset \mathbb{R}^k$.

\vspace{-0.3cm}
\paragraph{Continuous-time optimal control} 
In continuous time, we consider a dynamical system $\dot{s}(t)=f(s(t),a(t))$, where $s(t)$ denotes the state, $\dot{s}(t)$ its time derivative, and $a(t)\in\mathcal{A}$ the control action. Given an initial state $s(0)=s$ and a goal $g$, the undiscounted goal-reaching optimal control problem seeks a control trajectory $a(\cdot)$ that drives the system to $g$ while minimizing the cumulative cost $J = \int_0^T c(s(t),a(t))\,dt$, where $T$ denotes the goal-reaching time and $c(s(t),a(t))$ is the instantaneous cost. The optimal cost-to-go is defined as $d^*(s,g)=\inf_{a(\cdot)} \int_0^T c(s(t),a(t))\,dt$, with $s(0)=s$ and $s(T)=g$. By the Bellman principle of optimality, this cost satisfies $d^*(s,g) =\inf_{a\in\mathcal{A}} \left[c(s,a)\Delta t + d^*(s + f(s,a) \Delta t,g) \right] + o(\Delta t)$ for a short interval $\Delta t$. Under standard regularity conditions, expanding $d^*(s + f(s,a) \Delta t,g)$ to first order and taking the limit as $\Delta t\to 0$ yields
\begin{align}
    \begin{split}
        \inf_{a \in \mathcal{A}} [c(s, a) + \nabla_s d^*(s, g)^{\intercal}f(s,a)] = 0.
        \label{eq:HJB}
    \end{split}
\end{align}
Given $d^*(g,g)=0$, Eq.~\eqref{eq:HJB} represents the HJB PDE for the undiscounted goal-reaching problem.

\vspace{-0.3cm}
\paragraph{Physics-informed GCRL}
Pi-GCRL connects goal-conditioned value learning with continuous-time optimal control by augmenting standard discrete-time value losses with regularizers derived from PDEs. For example, an Eikonal-regularized implicit value-learning objective can be written as
\begin{align}
    \mathbb{E}_{\tau_\pi(g)}
    \Big[
        \underbrace{
        L_2^{\iota}
        \left(
            \mathcal{R}(s,g)
            +
            \gamma V_{\bar{\bm{\theta}}_V}(s',g)
            -
            V_{\bm{\theta}_V}(s,g)
        \right)
        }_{\text{value-learning loss}}
        +
        \underbrace{
        \left(
            \|\nabla_s V_{\bm{\theta}_V}(s,g)\| S(s)
            -
            1
        \right)^2
        }_{\text{Eikonal residual}}
    \Big],
    \label{eq:Eik_reg}
\end{align}
where $L_2^{\iota}$ denotes the expectile regression loss used in implicit value learning~\citep{park2024hiql}, and $\bar{\bm{\theta}}_V$ denotes target-network parameters. In \eqref{eq:Eik_reg}, the first term is a discrete-time value-learning loss, while the second penalizes violations of the Eikonal constraint, with $S(s)$ denoting a prescribed speed profile. This constraint is derived by the Eikonal PDE, $\|\nabla_s V_{\bm{\theta}_V}(s,g)\|^2 = 1/S(s)^2$, which classically describes wave propagation in heterogeneous media~\citep{noack2017acoustic}. Under the convention $V^*=-d^*$, prior work has shown that the Eikonal PDE and the HJB in~\eqref{eq:HJB} are closely related~\citep{giammarino2026physicsinformed, giammarino2026goal}. In particular, the two equations become equivalent when the dynamics are isotropic, i.e., when the feasible instantaneous velocity set $\mathcal{F}(s)=\{f(s,a):a\in\mathcal{A}\}$ is a scaled Euclidean ball for some $S(s)>0$. Eikonal-based regularization has proven effective in navigation and object-free goal-reaching. However, it becomes problematic in contact-rich tasks. We next formalize this failure mode.


\vspace{-0.1cm}
\section{Goal-Conditioned Value Learning under Hybrid Contact Dynamics}
\label{sec:analysis}

\vspace{-0.2cm}
This section analyzes the failure modes of Eikonal-based regularization for goal-conditioned value learning in contact-rich manipulation. We begin by modeling manipulation as a hybrid system~\citep{hedlund1999optimal}:
\begin{equation}
\begin{aligned}
    \dot{x}(t) = f_{q(t)}(x(t), z(t), u(t)), \ \ 
    \dot{z}(t) = g_{q(t)}(x(t), z(t), u(t)), \ \
    q(t) = \nu(x(t), z(t), q(t^-)),
\end{aligned}
\label{eq:hybrid_dynamics}
\end{equation}
where $x(t) \in \mathcal{X} \subset \mathbb{R}^n$ denotes the continuous state of the agent, $z(t) \in \mathcal{Z} \subset \mathbb{R}^m$ denotes the continuous state of objects in the environment, $q(t) \in \mathcal{Q} \subset \mathbb{Z}$ denotes the discrete contact mode, and $u(t) \in \mathcal{U}$ is the continuous control input. The map $\nu$ captures contact-dependent transitions, with $q(t^-)$ denoting the mode immediately before the transition. In contact-rich manipulation, the object dynamics $g_q$ depend strongly on the contact mode. For example, before contact, an object may be locally uncontrollable, so that $g_q(x,z,u)=0$, whereas during grasping or pushing, the same object coordinates may become controllable through their coupling with the agent dynamics. We consider manipulation tasks in which the goal involves controlling object states, such as pick-and-place. In this setting, we want to determine when Eikonal-based constraints in \eqref{eq:Eik_reg} fail to provide a sound inductive bias. Proposition~\ref{prop:eikonal_uncontrollable} answers this question by showing that full-state Eikonal regularization can be misaligned with locally uncontrollable directions of the hybrid dynamics in \eqref{eq:hybrid_dynamics}.

\begin{proposition}[Eikonal mismatch under locally uncontrollable coordinates]
\label{prop:eikonal_uncontrollable}
Consider a fixed mode $q$ of the hybrid system in \eqref{eq:hybrid_dynamics}, and suppose the state decomposes as $s=(x,z)$ with dynamics $\dot{x}=f_q(x,z,u)$ and $\dot{z}=0$. Here, $x$ denotes the locally controllable coordinates and $z$ denotes the locally uncontrollable coordinates in mode $q$. Assume unit running cost and let $d^*(x,z,g)$ be differentiable at $(x,z)$. If the locally uncontrollable coordinates affect the remaining cost-to-go, i.e., $\nabla_z d^*(x,z,g)\neq 0$, then, the full-state Eikonal constraint contradicts the local HJB characterization.
\vspace{-0.6cm}
\begin{proof}
    The HJB equation in this mode depends only on gradients along the controllable coordinates, $\min_{u\in\mathcal{U}}\left\{1+\nabla_x d^*(x,z,g)^\top f_q(x,z,u)\right\}=0$. To make the comparison most favorable to the Eikonal constraint, consider the isotropic case $f_q(x,z,u)=u$ with $\|u\|\leq 1$. Then, the HJB equation implies $\|\nabla_x d^*(x,z,g)\|=1$. However, the full-state Eikonal equation requires $\|\nabla_x d^*(x,z,g)\|^2 + \|\nabla_z d^*(x,z,g)\|^2=1$. Combining this condition with the local HJB $\|\nabla_x d^*(x,z,g)\|=1$ gives $1+\|\nabla_z d^*(x,z,g)\|^2=1$, which can hold only when $\nabla_z d^*(x,z,g)=0$. Hence, contradicting the initial assumption that $\nabla_z d^*(x,z,g)\neq 0$. In other words, whenever $d^*$ depends on locally uncontrollable coordinates, the full-state Eikonal constraint cannot be satisfied by the HJB solution: it incorrectly allocates part of the unit gradient budget to coordinates that are not locally controllable.
\end{proof}
\end{proposition}
\vspace{-0.3cm}
Proposition~\ref{prop:eikonal_uncontrollable} shows that the failure of Eikonal-based constraints in contact-rich manipulation is not due to nonsmooth value landscapes, but to a mismatch with the geometry of the dynamics. The optimal cost-to-go may vary along object coordinates $z$ that matter for long-horizon goal reaching, i.e., $\nabla_z d^*(x,z,g)\neq 0$, even though these coordinates are locally uncontrollable in the current contact mode, i.e., $\dot z=0$. The local HJB equation respects this structure by constraining value gradients only along feasible instantaneous motions. In contrast, the full-state Eikonal equation constrains the entire gradient and can therefore penalize valid value variation along uncontrollable coordinates.

This result further shows that Lipschitz continuity of the optimal value function, although useful in prior Eikonal-regularized GCRL settings~\citep{giammarino2026physicsinformed,giammarino2026goal}, is not sufficient in hybrid manipulation. PDE-based regularization should instead be imposed along the locally feasible directions of each contact mode. The contact-aware residual in Eq.~\eqref{eq:ca_hjb} implements this principle by replacing the full gradient-norm constraint in Eq.~\eqref{eq:Eik_reg} with a directional constraint along observed feasible transitions. Finally, we note that this failure mode is distinct from failures due to anisotropic dynamics. We refer the reader to the supplementary material for more on this distinction (Appendix~\ref{sec_app:analysis}).

\vspace{-0.2cm}
\section{Contact-Aware and Hierarchical Value Learning}
\label{sec:CA_and_Hierarchy}
\vspace{-0.2cm}

Guided by the analysis above, we introduce two formulations that align physics-informed regularization with locally feasible dynamics: $(i)$ a contact-aware residual that constrains value gradients along observed feasible transitions, and $(ii)$ a hierarchical decomposition that applies PDE-based regularization in a controllable low-level representation.

\vspace{-0.3cm}
\paragraph{Contact-aware physics-informed value learning} 
Consider a transition $(s,s')$ where $s$ includes both locally controllable and locally uncontrollable coordinates. To make the physics-informed constraint contact-aware, we approximate the direction of the continuous-time dynamics $f(s,a)$ in \eqref{eq:HJB} using the empirical transition direction $s'-s$. This yields the directional HJB-style residual
\begin{equation}
\mathcal{L}_{\mathrm{CA\text{-}HJB}}(\bm \theta)
=
\mathbb{E}_{\tau_\pi(g)}
\left[
\left(
\nabla_s d_{\bm \theta}(s,g)^\top
\frac{s'-s}{\|s'-s\|+\epsilon}
+1
\right)^2
\right],
\label{eq:ca_hjb}
\end{equation}
where $d_{\bm \theta}(s,g)=-V_{\bm \theta}(s,g)$ and $\epsilon>0$ is used for numerical stability. The normalization makes the residual depend on the observed direction of motion rather than on transition magnitude, which can vary with discretization, sampling rate, or state scaling. Unlike the full-state Eikonal residual in \eqref{eq:Eik_reg}, Eq.~\eqref{eq:ca_hjb} constrains the value gradient only along observed feasible motions, so locally uncontrollable coordinates do not contribute when their empirical displacement is zero. Section~\ref{sec:experiments} shows that this contact-aware modification improves performance in contact-rich manipulation.

\vspace{-0.3cm}
\paragraph{Physics-informed Hierarchical Flow}
We next propose a hierarchical formulation motivated by the same controllability principle. Instead of imposing PDE-based constraints on the full manipulation state, we separate high-level object-centric reasoning from low-level goal reaching, and apply the PDE regularizer only in a low-level representation aligned with the locally controllable dynamics.

Specifically, we define a high-level policy $\pi_{\mathrm{hi}} : \mathcal{S} \times \mathcal{G} \rightarrow \mathcal{X}$, which maps the full state $s \in \mathcal{S}$ and goal $g \in \mathcal{G}$ to a subgoal $x_g \in \mathcal{X}$. Here, $\mathcal{S}$ denotes the full manipulation state space, while $\mathcal{X}$ denotes the controllable state representation associated with the continuous state variable $x$ in \eqref{eq:hybrid_dynamics}. The low-level policy $\pi_{\mathrm{lo}} : \mathcal{X} \times \mathcal{X} \rightarrow \mathcal{A}$ then operates entirely within $\mathcal{X}$, mapping the current low-level state $x_s \in \mathcal{X}$ and subgoal $x_g \in \mathcal{X}$ to a primitive action. As a result, we obtain the following decomposition:
\[
(s,g)
\xrightarrow{\;\pi_{\mathrm{hi}}\;}
x_g,\qquad
(x_s,x_g)
\xrightarrow{\;\pi_{\mathrm{lo}}\;}
a.
\]
Built on this decomposition, we introduce \emph{Physics-informed Hierarchical Flow} (Pi-H-Flow), which combines a flow-based high-level subgoal generator~\citep{lyu2026flow} with low-level physics-informed value learning. Since the low-level learner acts over the controllable representation $\mathcal{X}$, Pi-GCRL objectives are better aligned with the local directions that the policy can influence. The low-level learner can therefore use any physics-informed objectives, including Eik-QRL~\citep{giammarino2026goal}, Eik-GCIVL as in \eqref{eq:Eik_reg}, or the contact-aware HJB-style residual in \eqref{eq:ca_hjb}.

We parameterize $\pi_{\mathrm{hi}}$ with a flow model because of its capacity to generate subgoals directly in the controllable representation, making them interpretable and avoiding an additional decoding step~\citep{park2026horizon}. To train the high-level flow, we sample subgoal targets from the replay buffer as future controllable states along the same trajectory. Specifically, for a sampled state $s$, we let $x_1$ be the controllable representation of a state reached after a fixed number of environment steps. Given $x_1$, we sample noise $x_0 \sim \mathcal{N}(0,I)$ and interpolation time $t\sim \mathcal{U}[0,1)$, define $x_t = (1-t)x_0 + t x_1$ with target velocity $y=x_1-x_0$, and train the flow field $v_{\bm \phi}$ with
\[
\mathcal{L}_{\mathrm{hi\text{-}flow}}(\bm \phi)
=
\mathbb{E}\left[
\left\|v_{\bm \phi}(s,g,x_t,t) - y\right\|^2
\right].
\]
We also train a high-level critic $Q_{\bm \psi}^{\mathrm{hi}}(s,g,x_g)$ using temporal-difference updates over temporally extended subgoal transitions. At evaluation time, subgoals are generated by sampling an initial noise vector $x_0 \sim \mathcal{N}(0,I)$ and integrating the learned flow field $v_{\bm \phi}$ from $t=0$ to $t=1$ to obtain a candidate subgoal $x_g$. Following~\citet{park2026horizon}, we sample multiple candidate subgoals from the learned flow, score them with the high-level critic, and pass the highest-valued subgoal to the low-level policy. We refer to the supplementary material for a summary of the full pipeline (Appendix~\ref{sec_app:algorithms_and_hyper}).

\section{Experiments}
\label{sec:experiments}

\begin{figure}
    \centering
    \includegraphics[width=\linewidth]{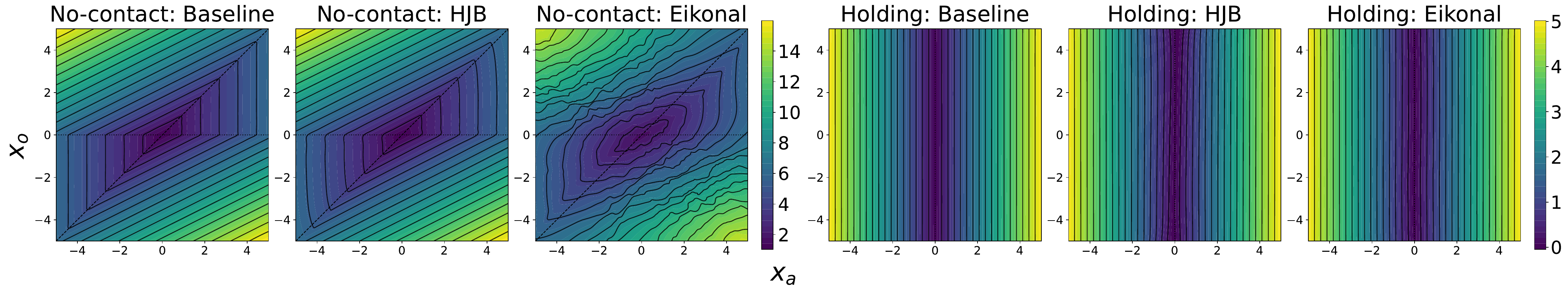}
    \caption{Results for regressing $d^*$ in \eqref{eq:toy_example_gc_value_function}. In the no-contact mode, full-state Eikonal regularization distorts the learned function $d_{\bm \theta}$ due to locally uncontrollable coordinates, as highlighted in Proposition~\ref{prop:eikonal_uncontrollable}. The HJB regularizer addresses this issue by constraining only the controllable direction. In the holding mode, where all coordinates are locally controllable, all losses recover similar solutions.}
    \label{fig:toy_example}
    \vspace{-0.3cm}
\end{figure}

\begin{wrapfigure}{r}{0.50\textwidth}
    \centering
    \begin{subfigure}[t]{0.15\textwidth}
        \centering
        \includegraphics[width=\linewidth]{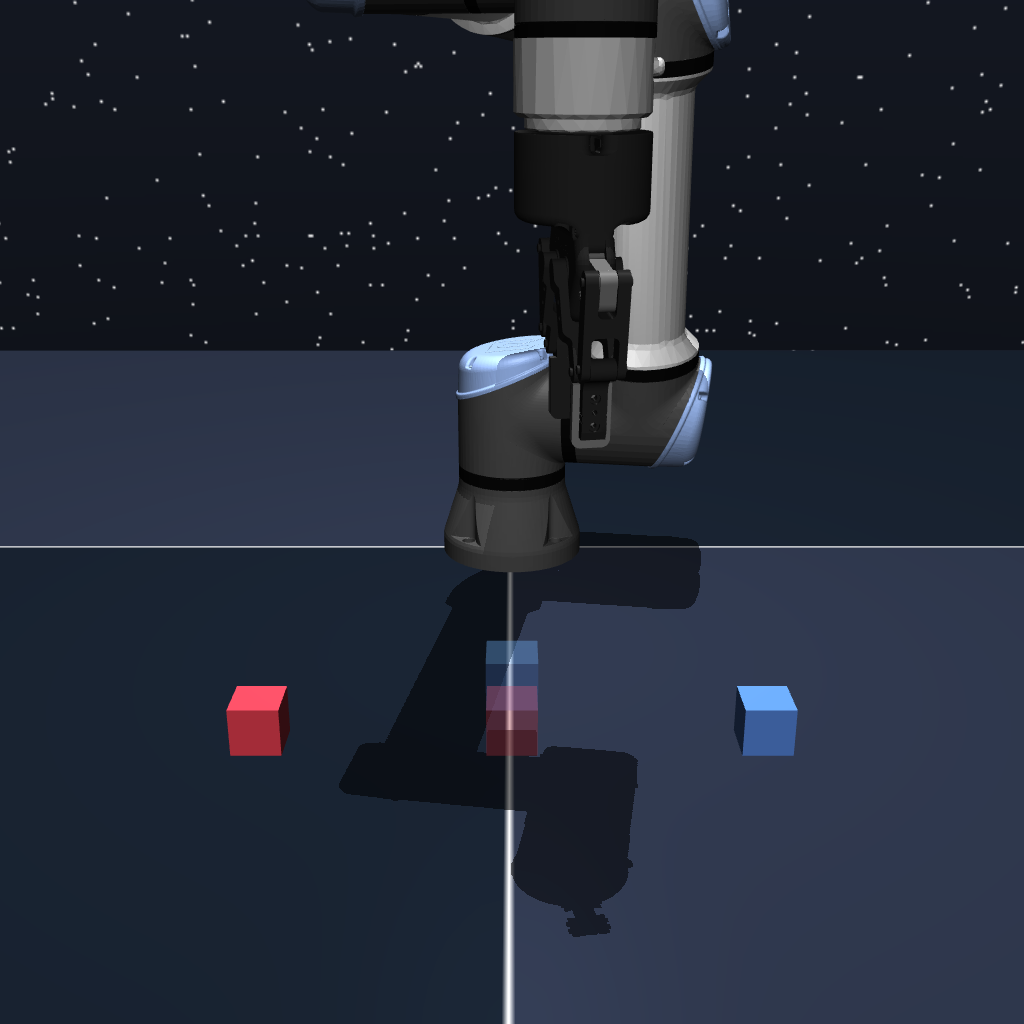}
        \caption{\texttt{Double}}
        \label{fig:Double}
    \end{subfigure}
    \begin{subfigure}[t]{0.15\textwidth}
        \centering
        \includegraphics[width=\linewidth]{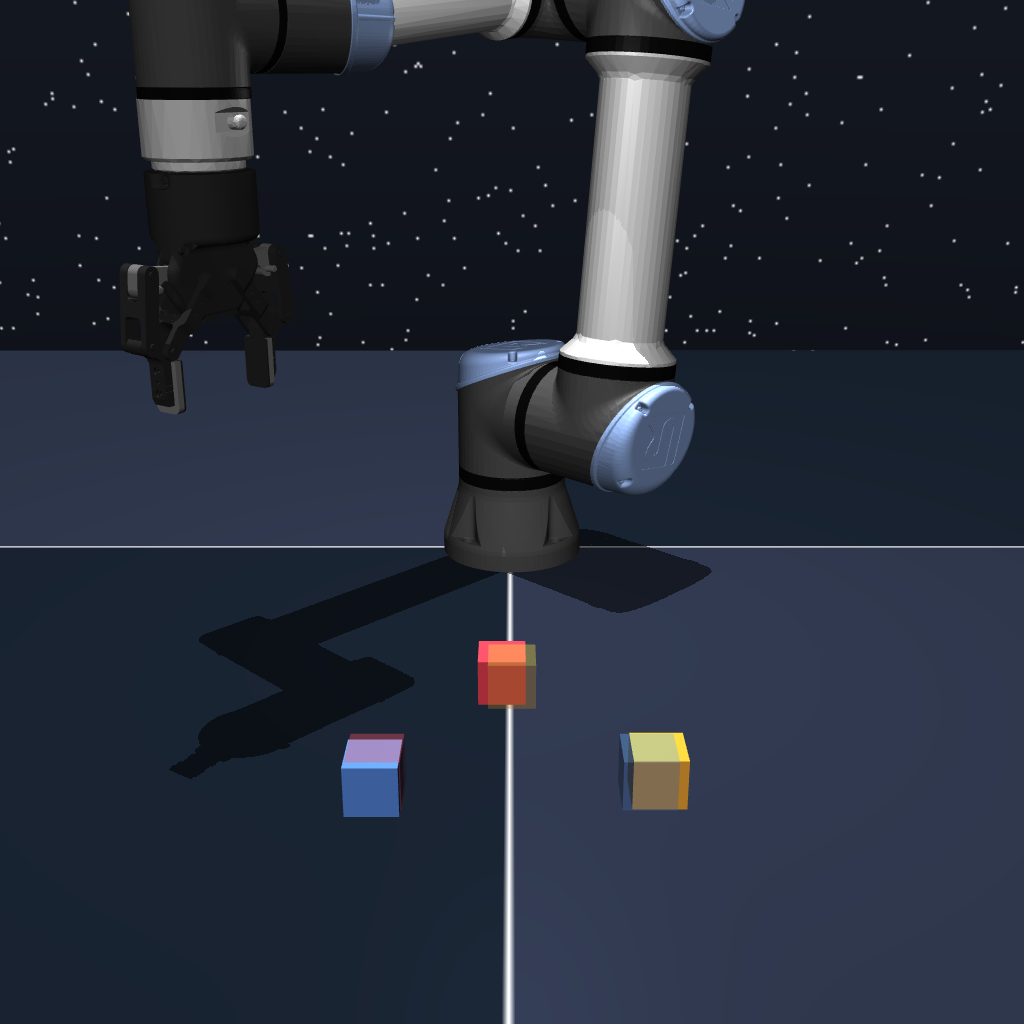}
        \caption{\texttt{Triple}}
        \label{fig:Triple}
    \end{subfigure}
    \begin{subfigure}[t]{0.15\textwidth}
        \centering
        \includegraphics[width=\linewidth]{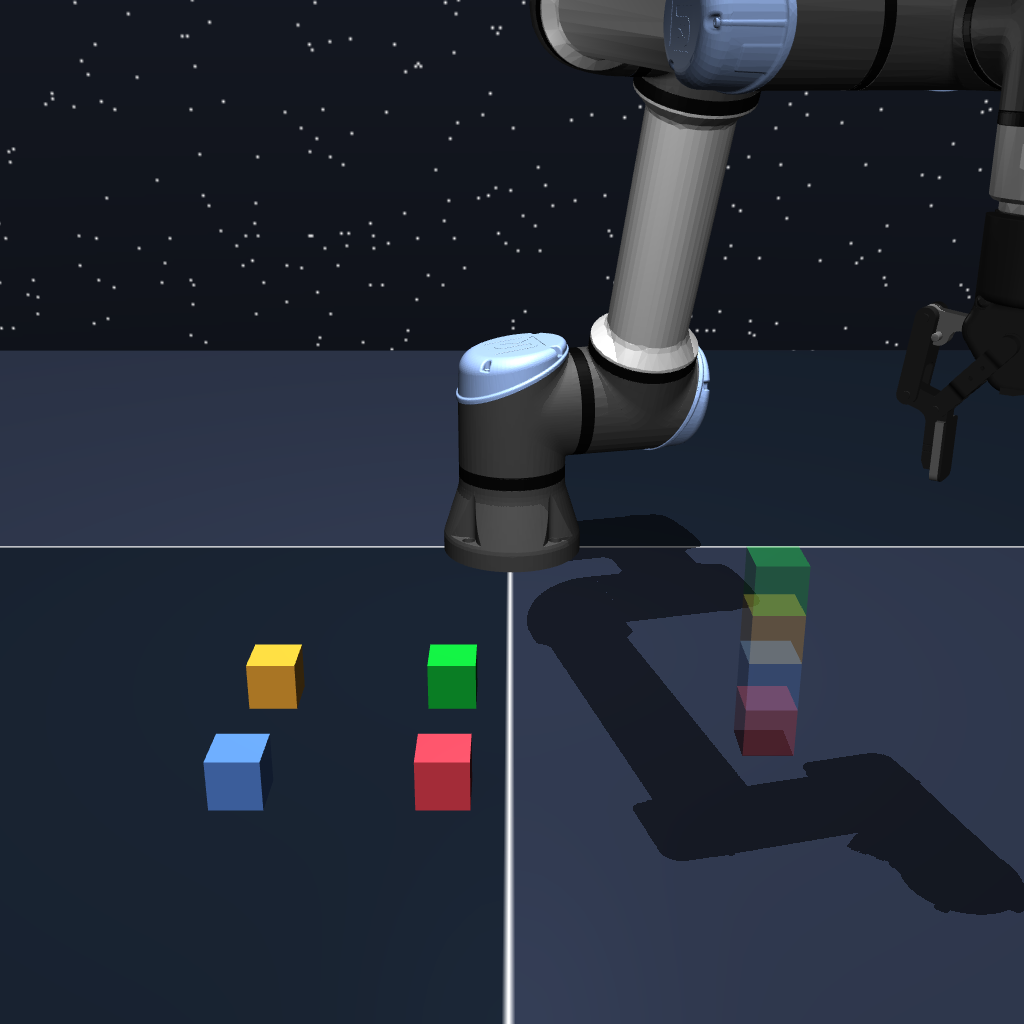}
        \caption{\texttt{Quadruple}}
        \label{fig:Quadruple}
    \end{subfigure}
    \vspace{-3pt}
    \caption{OGBench~\citep{park2025ogbench} manipulation environments used in our experiments. The state includes robot proprioception, end-effector pose, gripper state, and object poses, and the goal is to arrange the cubes into a specified target configuration.}
    \label{fig:cube_envs}
    \vspace{-10pt}
\end{wrapfigure}

\vspace{-0.3cm}
We organize our experiments around four main questions: $(1)$ does full-state Eikonal regularization distort value learning when some coordinates are locally uncontrollable? $(2)$ does this failure mode translate into degraded performance in contact-rich manipulation? $(3)$ does a controllable low-level representation improve hierarchical learning? and $(4)$ can the resulting framework be applied to real-world scenarios?

\vspace{-0.3cm}
\paragraph{A 1-D illustrative example}
We introduce a simple example to visualize the failure mode of full-state Eikonal regularization, as discussed in Proposition~\ref{prop:eikonal_uncontrollable}. Consider an MDP with $s=(x_a,x_o,h)$, where $x_a$ is the agent position, $x_o$ is the object position, and $h\in\{0,1\}$ indicates whether the agent is holding the object. The agent can move left or right, and can switch to the holding mode only when it is co-located with the object. Once $h=1$, the object moves together with the agent. For goals where the object must be held at the target location, i.e., $h_g=1$ and $x_a^g=x_o^g$, the undiscounted optimal cost-to-go under unit step cost is
\begin{equation}
    d^*(x_a, x_o, h; x_o^g, h_g=1) =
    \begin{cases}
        |x_a - x_o^g|, & h=1, \\[4pt]
        |x_a - x_o| + 1 + |x_o - x_o^g|, & h=0.
    \end{cases}
    \label{eq:toy_example_gc_value_function}
\end{equation}

Given $d^*$, we fix $x_o^g=0$ and train $d_{\bm \theta}$ with three different losses: $(i)$ an unregularized baseline
$\mathcal{L}(d_{\bm \theta}, d^*) = \mathbb{E}[(d_{\bm \theta}(s,g) - d^*(s,g))^2]$; $(ii)$ an HJB-regularized loss
$\mathcal{L}_{\mathrm{HJB}}(d_{\bm \theta}, d^*) = \mathcal{L}(d_{\bm \theta}, d^*) + \mathcal{L}_{\mathrm{CA\text{-}HJB}}(\bm{\theta})$, with $\mathcal{L}_{\mathrm{CA\text{-}HJB}}(\bm{\theta})$ defined in \eqref{eq:ca_hjb}; and $(iii)$ an Eikonal-regularized loss
$\mathcal{L}_{\mathrm{Eik}}(d_{\bm \theta}, d^*) = \mathcal{L}(d_{\bm \theta}, d^*) + \mathcal{L}_{\mathrm{Eik}}(\bm{\theta})$, where $\mathcal{L}_{\mathrm{Eik}}(\bm{\theta})$ is the Eikonal residual in \eqref{eq:Eik_reg}. The results are summarized in Fig.~\ref{fig:toy_example}, where we plot the learned $d_{\bm \theta}$ for both the no-contact mode ($h=0$) and the holding mode ($h=1$). In Fig.~\ref{fig:toy_example}, for the no-contact mode both the baseline and the HJB-regularized solution recover the piecewise-linear structure of $d^*$. In contrast, the Eikonal-regularized solution clearly distorts the value landscape by smoothing and reshaping its dependence on $x_o$. In the holding mode, where all state coordinates are locally controllable, the three losses recover similar solutions. These results empirically support the analysis in Section~\ref{sec:analysis}. We refer to Appendix~\ref{sec_app:training_curves} for more on this example.

\vspace{-0.3cm}
\paragraph{Simulated experiments}

In the following, we evaluate the impact of physics-informed regularization in goal-conditioned contact-rich manipulation tasks. Our goal is to assess whether the failure mode identified in Section~\ref{sec:analysis} translates into degraded performance, and whether the proposed contact-aware and hierarchical formulations can address this limitation. We conduct experiments on the manipulation suite of OGBench~\citep{park2025ogbench}, shown in Fig.~\ref{fig:cube_envs}. These environments require a 6-DoF UR5e manipulator to pick and rearrange multiple cubes into a specified goal configuration. The full MDP state contains robot proprioception, end-effector pose, gripper state, and object poses. Importantly, these tasks are goal-conditioned, i.e., algorithms are not evaluated only on a single task but on their ability to generalize across different goal configurations. 

\begin{figure}
    \centering
    \includegraphics[width=\linewidth]{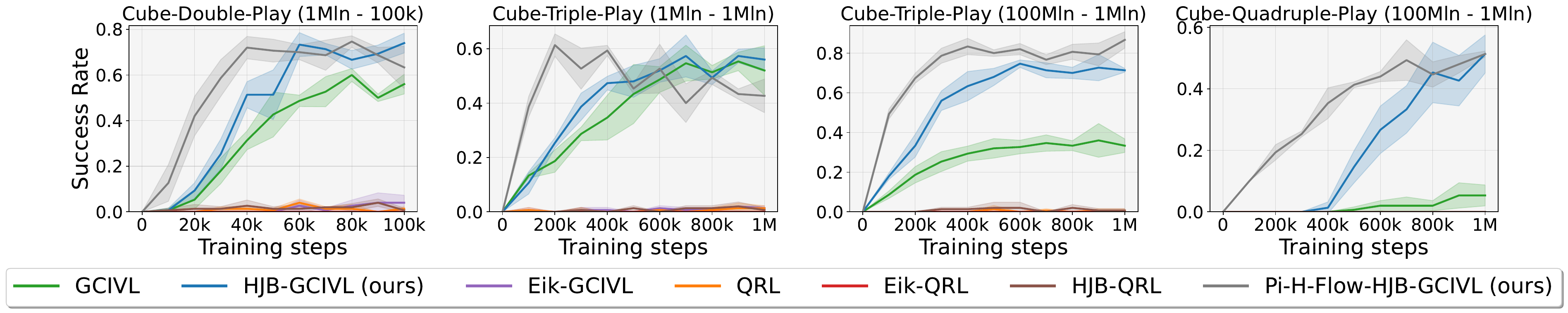}
    \caption{Results on the OGBench cube environments. The notation in each subplot title, e.g., \texttt{100Mln-1Mln}, indicates the dataset size and number of training steps, respectively. Evaluation is performed over $5$ goal configurations with $10$ episodes each and repeated over $3$ seeds. We report the mean and standard deviation across seeds.}
    \label{fig:ogbench_flat}
    \vspace{-10pt}
\end{figure}

We test our contact-aware formulation by replacing the full-state Eikonal residual in \eqref{eq:Eik_reg} with the HJB residual in \eqref{eq:ca_hjb}. We refer to this method as HJB-GCIVL. We also evaluate our hierarchical formulation, Pi-H-Flow, using HJB-GCIVL as the low-level learner; we refer to this variant as Pi-H-Flow-HJB-GCIVL. In Pi-H-Flow-HJB-GCIVL, the low-level learner operates on a controllable representation containing only manipulator-related state variables, excluding object-pose coordinates. We compare against GCIVL~\citep{park2025ogbench}, Eik-GCIVL~\citep{giammarino2026physicsinformed}, QRL~\citep{wang2023optimal}, and Pi-QRL variants, including Eik-QRL and HJB-QRL~\citep{giammarino2026goal}. The results are summarized in Fig.~\ref{fig:ogbench_flat}. QRL-based methods struggle in this setting, highlighting the rigidity of quasimetric parameterizations in high-dimensional manipulation state spaces, as also observed in prior work~\citep{giammarino2026goal}. Consistent with our analysis, Eik-GCIVL fails to converge reliably, reflecting the misalignment between full-state Eikonal regularization and contact-dependent controllability. In contrast, GCIVL learns meaningful policies, but both our proposed variants achieve substantially better performance, especially in the more challenging \texttt{Triple} and \texttt{Quadruple} environments. Finally, Pi-H-Flow-HJB-GCIVL outperforms and converges faster than HJB-GCIVL, supporting the benefit of hierarchy for long-horizon goal-conditioned manipulation.

\begin{figure}
    \centering
    \includegraphics[width=\linewidth]{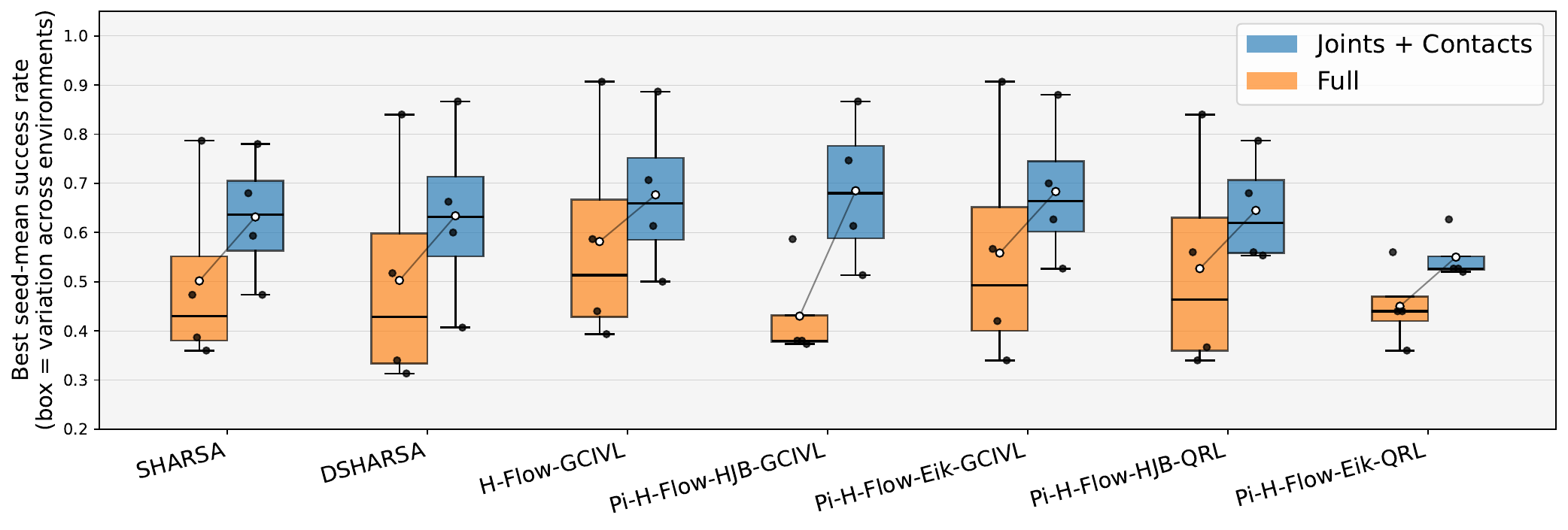}
    \caption{State representation ablation for hierarchical algorithms. For each algorithm, representation, and environment, we first average the evaluation curves across seeds and then take the best checkpoint of the resulting mean curve. Evaluation is performed as in Fig.~\ref{fig:ogbench_flat}. Each box summarizes these best seed-mean success rates across environments. Black dots denote individual environments, while white dots denote the mean across environments. We report the full training curves in Appendix~\ref{sec_app:training_curves}.}
    \label{fig:ogbench_hierarchical}
    \vspace{-15pt}
\end{figure}

\vspace{-0.3cm}
\paragraph{State representation ablation.}
An important question is whether the proposed state decomposition is actually useful for hierarchical algorithms. Pi-H-Flow is designed to use different representations at the high and low levels. However, one could argue that, with appropriately defined subgoals, using the full manipulation state or a controllable low-level representation should lead to similar performance, since the subgoals may already restrict the low-level problem to controllable variables or modes. We test this hypothesis using the same environments, tasks, and evaluation protocol as in Fig.~\ref{fig:ogbench_flat}.

Results are summarized in Fig.~\ref{fig:ogbench_hierarchical}, where we compare hierarchical algorithms whose low-level MDP is defined either on the full manipulation state or on the proposed controllable representation, denoted as ``Joints+Contacts'', which excludes object-pose coordinates. We also include SHARSA and DSHARSA~\citep{park2026horizon} as hierarchical baselines. Across algorithms, using the controllable low-level representation consistently improves performance over using the full state, with gains between $10\%-20\%$ in best mean success rate across seeds for most methods. This improvement is not limited to physics-informed objectives, suggesting that goal reaching in controllable robot-centric coordinates is an easier low-level problem than goal reaching in the full object-centric manipulation state. For physics-informed methods, the same decomposition further aligns the PDE regularizer with the locally feasible directions of the low-level dynamics. Among the evaluated variants, Pi-H-Flow-HJB-GCIVL achieves the strongest overall performance, supporting the combination of contact-aware regularization with the proposed hierarchical MDP decomposition.

\begin{wrapfigure}{r}{0.5\textwidth}
    \centering
    \vspace{-12pt}
    \includegraphics[width=\linewidth]{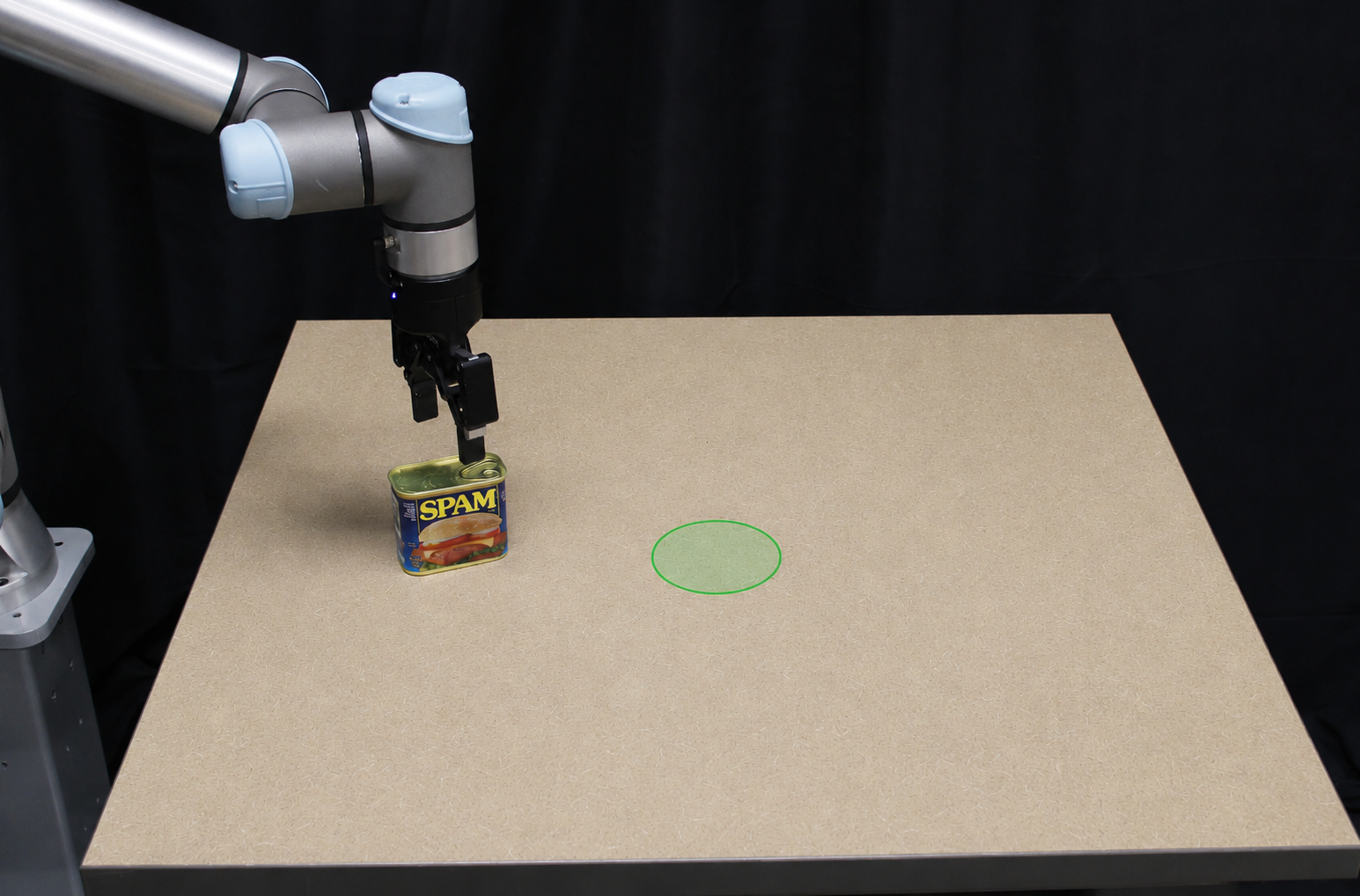}
    \caption{Real-world experimental setup. The task consists of moving an object to the center of the table from randomized initial poses.}
    \label{fig:setup_real_world_experiments}
    \vspace{-8pt}
\end{wrapfigure}

\vspace{-0.3cm}
\paragraph{Real-world experiments.}
We further evaluate the proposed framework on the real-world pick-and-place task shown in Fig.~\ref{fig:setup_real_world_experiments}. Due to space constraints, we summarize the main findings here and provide additional details in Appendix~\ref{sec_app:real_world_experiments}. Given a randomized initial object pose, the robot must place the object at the center of the table. State and action spaces are defined analogously to the OGBench manipulation environments~\citep{park2025ogbench}, with object-pose estimates obtained from an RGB-D camera. For these experiments, we construct an offline dataset using a scripted data-collection policy based on an MPC controller~\citep{manganaris2026graph}. We collect approximately $60$k interaction steps and train four policies for $300$k gradient steps: GCIVL, HJB-GCIVL, H-Flow-GCIVL, and Pi-H-Flow-HJB-GCIVL. We then evaluate the trained policies on the physical system over $20$ episodes under three object-position thresholds, corresponding to $3$cm, $5$cm, and $10$cm Euclidean distance from the center of the table.

GCIVL achieves success rates of $25\%$, $40\%$, and $45\%$ at the three thresholds, respectively, while HJB-GCIVL improves to $40\%$, $50\%$, and $65\%$. H-Flow-GCIVL obtains $25\%$, $40\%$, and $40\%$, whereas Pi-H-Flow-HJB-GCIVL achieves the strongest performance, with success rates of $50\%$, $60\%$, and $75\%$. These results show that, even with a limited dataset of only $60$k real-world samples, all methods learn meaningful behavior from offline data, while the physics-informed variants generally improve over their non-physics-informed counterparts. These experiments showcase that the proposed controllability-aware decomposition and HJB-style regularization remain effective beyond simulation and are compatible with practical robot-learning pipelines. 

\vspace{-0.3cm}
\section{Conclusion}
\vspace{-0.3cm}

In this paper, we extended and analyzed Pi-GCRL for contact-rich manipulation. We showed that full-state Eikonal regularization can be misaligned with contact-dependent controllability, distorting value functions and degrading performance. This analysis motivated two complementary solutions: a contact-aware HJB-style regularizer that constrains value gradients along controllable directions, and a hierarchical decomposition that applies physics-informed regularization only in a low-level controllable representation. Across illustrative, simulated, and real-world experiments, these formulations improve over full-state regularization, especially in challenging multi-object manipulation tasks.

\vspace{-0.3cm}
\paragraph{Limitations and future work.}
While these results are encouraging, important challenges remain. First, all evaluated algorithms are still data hungry, as is common in GCRL. Although physics-informed objectives improve performance, stronger physics priors are needed to improve sample efficiency, especially during real-world training. Second, our HJB regularizer shapes value learning according to feasible directions of motion, but it does not by itself guarantee that the resulting policy avoids unsafe contacts, respects actuator or force limits, or adapts its impedance to object properties. Future work will incorporate more explicit physics models into the HJB formulation, rather than relying only on data-driven approximations of feasible transitions. We also plan to extend Pi-GCRL toward safety-aware and compliant manipulation, for example by integrating reachability, constraint satisfaction, contact-force-aware objectives, or impedance-aware value regularization. More broadly, we are interested in using Pi-GCRL as a structured approach for adapting pretrained vision-language-action models. This direction could connect semantic goals to low-level behaviors that encode properties such as fragility, material constraints, or preferred modes of interaction.


\bibliography{example}  

\clearpage
\appendix

\section{Ethical Statement}

This work is primarily methodological and studies physics-informed value learning for GCRL under hybrid contact dynamics. We do not identify direct application-specific risks beyond those generally associated with deploying learned robotic manipulation systems. Since contact-rich manipulation may be used in manufacturing, logistics, and automation, practical deployment would require appropriate safety validation and safeguards for operation near humans. The experiments in this paper are conducted in controlled simulated and laboratory settings, and the proposed methods are not intended to provide formal safety guarantees by themselves.

\section{Reproducibility Statement}

To promote transparency and reproducibility, we will release the code used for our experiments in the camera-ready version. The appendix further reports implementation details, hyperparameters, additional training curves, and details of the real-world experimental setup and evaluation protocol.

\section{LLM Usage Disclosure}

The authors used LLMs to support manuscript preparation, including text polishing, proofreading, and improving the clarity and rigor of theoretical and mathematical formulations. All LLM-assisted content was thoroughly reviewed by the authors, who assume full responsibility for the final manuscript.

\section{Analysis: local uncontrollability vs anisotropy}
\label{sec_app:analysis}

We now provide the second part of our theoretical analysis, which was omitted from the main text due to space constraints. Proposition~\ref{prop:eikonal_uncontrollable} shows that full-state Eikonal regularization can be incompatible with the local HJB characterization when some goal-relevant coordinates are locally uncontrollable. A natural question is whether this failure simply reflects the use of an isotropic Eikonal constraint for dynamics with anisotropic geometry. We argue that these are distinct issues. Under full-dimensional anisotropic dynamics, all state directions remain locally feasible, but with direction-dependent speeds or costs. In this case, the HJB equation in \eqref{eq:HJB} captures the correct control-induced geometry, while the Eikonal equation in \eqref{eq:Eik_reg} provides an isotropic approximation. In contrast, under local uncontrollability, the Eikonal constraint regularizes directions that are not locally feasible at all. The next proposition formalizes this distinction.

\begin{proposition}[Eikonal approximation under full-dimensional anisotropic dynamics]
\label{prop:eikonal_fully_controllable}
Consider a continuous-time goal-reaching problem with dynamics $\dot{s}=f(s,u)$, unit running cost, and differentiable optimal cost-to-go $d^*(s,g)$. Let $\mathcal{F}(s)=\{f(s,u):u\in\mathcal{U}\}$ denote the feasible instantaneous velocity set. The HJB equation characterizes the local goal-reaching geometry through the support function of $\mathcal{F}(s)$. If $\mathcal{F}(s)$ is the Euclidean unit ball $\mathbb{B}$, this characterization reduces to the Eikonal equation $\|\nabla_s d^*(s,g)\|=1$. If instead $\mathcal{F}(s)$ is full-dimensional but anisotropic, the Euclidean Eikonal equation can be interpreted as an isotropic approximation of the HJB-induced local geometry.
\vspace{-0.7cm}
\begin{proof}
With unit running cost, the HJB equation is
$
\min_{u\in\mathcal{U}}\{1+\nabla_s d^*(s,g)^\top f(s,u)\}=0.
$
Using $\mathcal{F}(s)=\{f(s,u):u\in\mathcal{U}\}$, this becomes
$
1+\min_{v\in\mathcal{F}(s)}\nabla_s d^*(s,g)^\top v=0.
$
Equivalently,
$
\sigma_{\mathcal{F}(s)}(-\nabla_s d^*(s,g))=1,
$
where $\sigma_{\mathcal{F}(s)}(p)=\sup_{v\in\mathcal{F}(s)}p^\top v$ is the support function of $\mathcal{F}(s)$. If $\mathcal{F}(s)=\mathbb{B}$, then $\sigma_{\mathcal{F}(s)}(p)=\|p\|$, yielding $\|\nabla_s d^*(s,g)\|=1$. When $\mathcal{F}(s)$ is full-dimensional but anisotropic, the support function still constrains gradients over all state directions, but according to the anisotropic feasible velocity set; replacing it with the Euclidean norm therefore gives an isotropic approximation.
\end{proof}
\end{proposition}

\vspace{-0.4cm}
Together, Proposition~\ref{prop:eikonal_uncontrollable} and Proposition~\ref{prop:eikonal_fully_controllable} lead to two main observations. First, full-state Eikonal regularization introduces a \textit{structural misalignment} when the dynamics contain locally uncontrollable coordinates, such as object states in no-contact modes. In this case, the Eikonal penalty constrains value gradients along directions that do not correspond to feasible instantaneous motion, as shown in Proposition~\ref{prop:eikonal_uncontrollable}. Second, Proposition~\ref{prop:eikonal_fully_controllable} clarifies why Eikonal regularization can still be effective in navigation and object-free goal-reaching tasks, even when the true dynamics are not perfectly isotropic. In these settings, goal-relevant directions are often locally feasible, at least approximately, so the Eikonal constraint remains an imperfect but dynamically meaningful approximation of the HJB geometry. This distinction further motivates the contact-aware and hierarchical formulations introduced in Section~\ref{sec:CA_and_Hierarchy} in the main text, which avoid imposing a single full-state Eikonal geometry across the entire manipulation problem.

\section{Algorithms and Implementation details}
\label{sec_app:algorithms_and_hyper}

\subsection{Training and inference pipeline for Pi-H-Flow-HJB-GCIVL}

We denote the full manipulation state by $s$, the low-level controllable representation by $x$, the task goal by $g$, and a high-level subgoal by $x_g$. Pi-H-Flow-HJB-GCIVL consists of a high-level value function $V^{\mathrm{hi}}_{\bm \theta}$, a double high-level critic $Q^{\mathrm{hi}}_{\bm \psi_1},Q^{\mathrm{hi}}_{\bm \psi_2}$, a high-level flow actor $v_{\bm \phi}$, a double low-level value function $V^{\mathrm{lo}}_{\bm \omega_1},V^{\mathrm{lo}}_{\bm \omega_2}$, and a low-level actor $\pi^{\mathrm{lo}}_{\bm \eta}$. Target-network parameters are denoted by adding a bar to the corresponding parameter vector, following the notation used in Eq.~\eqref{eq:Eik_reg} in the main text. We first summarize the main loss functions used to train these components, before presenting the full training and inference pipelines. These pipelines are respectively summarized in Algorithm~\ref{alg:pihflow_hjb_gcivl_train} and Algorithm~\ref{alg:pihflow_hjb_gcivl_inference}. For additional implementation details, including the dataset processing pipeline, model architectures, loss implementations, and training scripts, we refer to our GitHub repository.

\paragraph{High-level critic and value losses.}
The high-level critic is trained with temporally extended temporal-difference (TD) targets~\citep{sutton2018reinforcement}. For a subgoal transition of length $k$, we define
\begin{align}
    y^{\mathrm{hi}} 
    =
    \mathcal{R}^{\mathrm{hi}} 
    + \gamma^k m^{\mathrm{hi}} V^{\mathrm{hi}}_{\bm \theta}(s_{t+k},g),
\end{align}
where $m^{\mathrm{hi}}$ is the high-level continuation mask. The high-level critic loss is
\begin{align}
    \mathcal{L}_{Q}^{\mathrm{hi}}(\bm \psi)
    =
    \mathbb{E}_{\mathcal{D}}
    \left[
    \ell\!\left(Q^{\mathrm{hi}}_{\bm \psi_1}(s_t,g,x_g), y^{\mathrm{hi}}\right)
    +
    \ell\!\left(Q^{\mathrm{hi}}_{\bm \psi_2}(s_t,g,x_g), y^{\mathrm{hi}}\right)
    \right],
    \label{eq_app:hi_critic_loss}
\end{align}
where $\ell$ is the value loss used in implementation, e.g., binary cross-entropy for the reported experiments, and $\mathcal{D}$ is a replay buffer. The high-level value function is trained from the target high-level critic:
\begin{align}
    \mathcal{L}_{V}^{\mathrm{hi}}(\bm \theta)
    =
    \mathbb{E}_{\mathcal{D}}
    \left[
    \ell\!\left(
    V^{\mathrm{hi}}_{\bm \theta}(s_t,g),
    \min_{i=1,2} Q^{\mathrm{hi}}_{\bar{\bm \psi}_i}(s_t,g,x_g)
    \right)
    \right].
    \label{eq_app:hi_value_loss}
\end{align}

\paragraph{Low-level HJB-regularized value loss.}
The low-level value function operates on the controllable representation $x$. Let
\begin{align}
    y_i^{\mathrm{lo}}
    =
    \mathcal{R}^{\mathrm{lo}}
    +
    \gamma_{\mathrm{lo}} m^{\mathrm{lo}}
    V^{\mathrm{lo}}_{\bar{\bm \omega}_i}(x_{t+1},x_g),
    \qquad i \in \{1,2\},
\end{align}
and
\begin{align}
    y^{\mathrm{lo}}
    =
    \mathcal{R}^{\mathrm{lo}}
    +
    \gamma_{\mathrm{lo}} m^{\mathrm{lo}}
    \min_{i=1,2} V^{\mathrm{lo}}_{\bar{\bm \omega}_i}(x_{t+1},x_g).
\end{align}
The expectile advantage is computed using the target value estimate,
\begin{align}
    A^{\mathrm{lo}}
    =
    y^{\mathrm{lo}}
    -
    \frac{1}{2}
    \sum_{i=1}^{2}
    V^{\mathrm{lo}}_{\bar{\bm \omega}_i}(x_t,x_g).
\end{align}
The implicit value-learning term is
\begin{align}
    \mathcal{L}_{\mathrm{IVL}}^{\mathrm{lo}}(\bm \omega)
    =
    \sum_{i=1}^{2}
    \mathbb{E}_{\mathcal{D}}
    \left[
    L_2^{\iota}
    \left(
    A^{\mathrm{lo}},
    y_i^{\mathrm{lo}} - V^{\mathrm{lo}}_{\bm \omega_i}(x_t,x_g)
    \right)
    \right].
    \label{eq_app:low_ivl_loss}
\end{align}
We add the contact-aware HJB-style residual from the main text. Defining $d_{\bm \omega_i}(x,x_g)=-V^{\mathrm{lo}}_{\bm \omega_i}(x,x_g)$ and
\begin{align}
    u_t = \frac{x_{t+1}-x_t}{\|x_{t+1}-x_t\|+\epsilon},
\end{align}
the low-level HJB residual is
\begin{align}
    \mathcal{L}_{\mathrm{HJB}}^{\mathrm{lo}}(\bm \omega)
    =
    \sum_{i=1}^{2}
    \mathbb{E}_{\mathcal{D}}
    \left[
    \left(
    \nabla_x d_{\bm \omega_i}(x_t,x_g)^\top u_t + 1
    \right)^2
    \right].
    \label{eq_app:low_hjb_loss}
\end{align}
The complete low-level value loss is
\begin{align}
    \mathcal{L}_{V}^{\mathrm{lo}}(\bm \omega)
    =
    \mathcal{L}_{\mathrm{IVL}}^{\mathrm{lo}}(\bm \omega)
    +
    \lambda_{\mathrm{HJB}}
    \mathcal{L}_{\mathrm{HJB}}^{\mathrm{lo}}(\bm \omega),
    \label{eq_app:low_value_loss}
\end{align}
with $\lambda_{\mathrm{HJB}}=1$ in our implementation.

\paragraph{High-level flow actor loss.}
The high-level actor generates subgoals in the controllable representation. Given a future controllable state $x_1$ sampled from the same trajectory, we sample $x_0 \sim \mathcal{N}(0,I)$ and $t\sim \mathcal{U}[0,1]$, define
\begin{align}
    x_t = (1-t)x_0 + t x_1,
    \qquad
    y = x_1 - x_0,
\end{align}
and train the high-level flow field with
\begin{align}
    \mathcal{L}_{\mathrm{flow}}^{\mathrm{hi}}(\bm \phi)
    =
    \mathbb{E}_{\mathcal{D},x_0,t}
    \left[
    \left\|
    v_{\bm \phi}(s,g,x_t,t)-y
    \right\|^2
    \right].
    \label{eq_app:hi_flow_loss}
\end{align}

\paragraph{Low-level actor loss.}
The low-level actor is trained with advantage-weighted regression. We define
\begin{align}
    A^{\mathrm{actor}}_{\mathrm{lo}}
    =
    \frac{1}{2}\sum_{i=1}^{2}
    V^{\mathrm{lo}}_{\bm \omega_i}(x_{t+1}, x_g)
    -
    \frac{1}{2}\sum_{i=1}^{2}
    V^{\mathrm{lo}}_{\bm \omega_i}(x_t, x_g).
\end{align}
The low-level actor loss is
\begin{align}
    \mathcal{L}_{\pi}^{\mathrm{lo}}(\bm \eta)
    =
    -
    \mathbb{E}_{\mathcal{D}}
    \left[
    \min\left(\exp(\alpha A^{\mathrm{actor}}_{\mathrm{lo}}),100\right)
    \log \pi^{\mathrm{lo}}_{\bm \eta}(a_t|x_t,x_g)
    \right].
    \label{eq_app:low_actor_loss}
\end{align}

\begin{algorithm}[H]
\caption{Training Pi-H-Flow-HJB-GCIVL}
\label{alg:pihflow_hjb_gcivl_train}
\begin{algorithmic}
\STATE {\bfseries Input:} Offline dataset $\mathcal{D}$, high-level value $V^{\mathrm{hi}}_{\bm \theta}$, high-level critic $Q^{\mathrm{hi}}_{\bm \psi}$, high-level flow actor $v_{\bm \phi}$, low-level value $V^{\mathrm{lo}}_{\bm \omega}$, low-level actor $\pi^{\mathrm{lo}}_{\bm \eta}$, target networks, target update rate $\tau$
\WHILE{not converged}
    \STATE Sample a batch from $\mathcal{D}$
    \STATE Parse full states $s$ and controllable states $x$
    \STATE Update $V^{\mathrm{hi}}_{\bm \theta}$ using Eq.~\eqref{eq_app:hi_value_loss}
    \STATE Update $Q^{\mathrm{hi}}_{\bm \psi}$ using Eq.~\eqref{eq_app:hi_critic_loss}
    \STATE Update $V^{\mathrm{lo}}_{\bm \omega}$ using Eq.~\eqref{eq_app:low_value_loss}
    \STATE Update $v_{\bm \phi}$ using Eq.~\eqref{eq_app:hi_flow_loss}
    \STATE Update $\pi^{\mathrm{lo}}_{\bm \eta}$ using Eq.~\eqref{eq_app:low_actor_loss}
    \STATE Update target networks:
    $
    \bar{\bm \psi} \leftarrow (1-\tau)\bar{\bm \psi}+\tau\bm\psi,
    \quad
    \bar{\bm \omega} \leftarrow (1-\tau)\bar{\bm \omega}+\tau\bm\omega
    $
\ENDWHILE
\end{algorithmic}
\end{algorithm}

\begin{algorithm}[H]
\caption{Inference with Pi-H-Flow-HJB-GCIVL}
\label{alg:pihflow_hjb_gcivl_inference}
\begin{algorithmic}
\STATE {\bfseries Input:} Current full state $s$, current controllable state $x$, goal $g$, high-level flow actor $v_{\bm \phi}$, high-level critic $Q^{\mathrm{hi}}_{\bm \psi}$, low-level actor $\pi^{\mathrm{lo}}_{\bm \eta}$, number of samples $N$, flow steps $M$
\STATE Sample $N$ initial subgoals $x_0^i \sim \mathcal{N}(0,I)$
\FOR{$m=0,\ldots,M-1$}
    \STATE Set $t_m=m/M$
    \STATE Update each candidate:
    $
    x_{m+1}^i = x_m^i + \frac{1}{M}v_{\bm \phi}(s,g,x_m^i,t_m)
    $
\ENDFOR
\STATE Score candidates using the high-level critic:
$
q_i = \min_{j=1,2} Q^{\mathrm{hi}}_{\bm \psi_j}(s,g,x_M^i)
$
\STATE Select the best subgoal:
$
x_g = x_M^{\arg\max_i q_i}
$
\STATE Sample low-level action:
$
a \sim \pi^{\mathrm{lo}}_{\bm \eta}(\cdot|x,x_g)
$
\STATE Clip $a$ to the action bounds
\STATE \textbf{return} $a$
\end{algorithmic}
\end{algorithm}

\subsection{Hyperparameters}

All methods use the hyperparameters reported in Table~\ref{tab:pihflow_hyperparameters}. For Pi-H-Flow-HJB-GCIVL, the high-level flow actor uses $M=10$ integration steps at inference and samples $N=32$ candidate subgoals, which are ranked using the high-level critic. The low-level discount is set automatically as $1 - 1/K$, where $K$ is the number of subgoal steps.

\begin{table}[H]
\centering
\caption{Hyperparameters used for Pi-H-Flow-HJB-GCIVL.}
\label{tab:pihflow_hyperparameters}
\begin{tabular}{l c}
\toprule
\textbf{Hyperparameter} & \textbf{Value} \\
\midrule
Learning rate & $3 \times 10^{-4}$ \\
Batch size & $1024$ \\
Actor hidden dimensions & $(1024,1024,1024,1024)$ \\
Value hidden dimensions & $(1024,1024,1024,1024)$ \\
Expectile $\iota$ & $0.9$ \\
AWR temperature $\alpha$ & $10.0$ \\
Layer normalization & True \\
Constant actor standard deviation & True \\
Discount factor $\gamma$ & $0.999$ \\
Low-level discount $\gamma_{\mathrm{lo}}$ & $1 - 1/K$ \\
Target update rate $\tau$ & $0.005$ \\
Flow integration steps & $10$ \\
Candidate subgoals at inference & $32$ \\
Low-level observation type & Joints+Contacts \\
Subgoal horizon $K$ & $25$ \\
Value goal: current state probability & $0.2$ \\
Value goal: trajectory future probability & $0.5$ \\
Value goal: random state probability & $0.3$ \\
Actor goal: current state probability & $0.0$ \\
Actor goal: trajectory future probability & $1.0$ \\
Actor goal: random state probability & $0.0$ \\

\bottomrule
\end{tabular}
\end{table}

\section{Experiments: additional details and training curves}
\label{sec_app:training_curves}

\subsection{Computational resources}
All experiments follow a standard offline RL pipeline, where datasets are collected before training and algorithms are trained only on fixed offline data. Training was performed on a workstation equipped with four NVIDIA GeForce RTX 3090 GPUs, each with 24GB of memory. We parallelized independent training runs across GPUs, with each run using a single GPU. In our implementation, the GPU memory usage of each algorithm remained below approximately 4GB, and training for one million gradient steps required approximately five hours per run.

\subsection{Extended analysis of the hybrid toy example}

In Section~\ref{sec:experiments}, we used a simple hybrid manipulation example to illustrate the failure mode predicted by Proposition~\ref{prop:eikonal_uncontrollable}. Here, we provide additional details on the construction of the example, the closed-form optimal cost-to-go, and the regression losses used to generate Fig.~\ref{fig:toy_example}. The goal is to isolate the role of local uncontrollability from other sources of approximation error and to show explicitly how full-state Eikonal regularization behaves differently across contact modes.

\begin{figure}
    \centering
    \includegraphics[width=\linewidth]{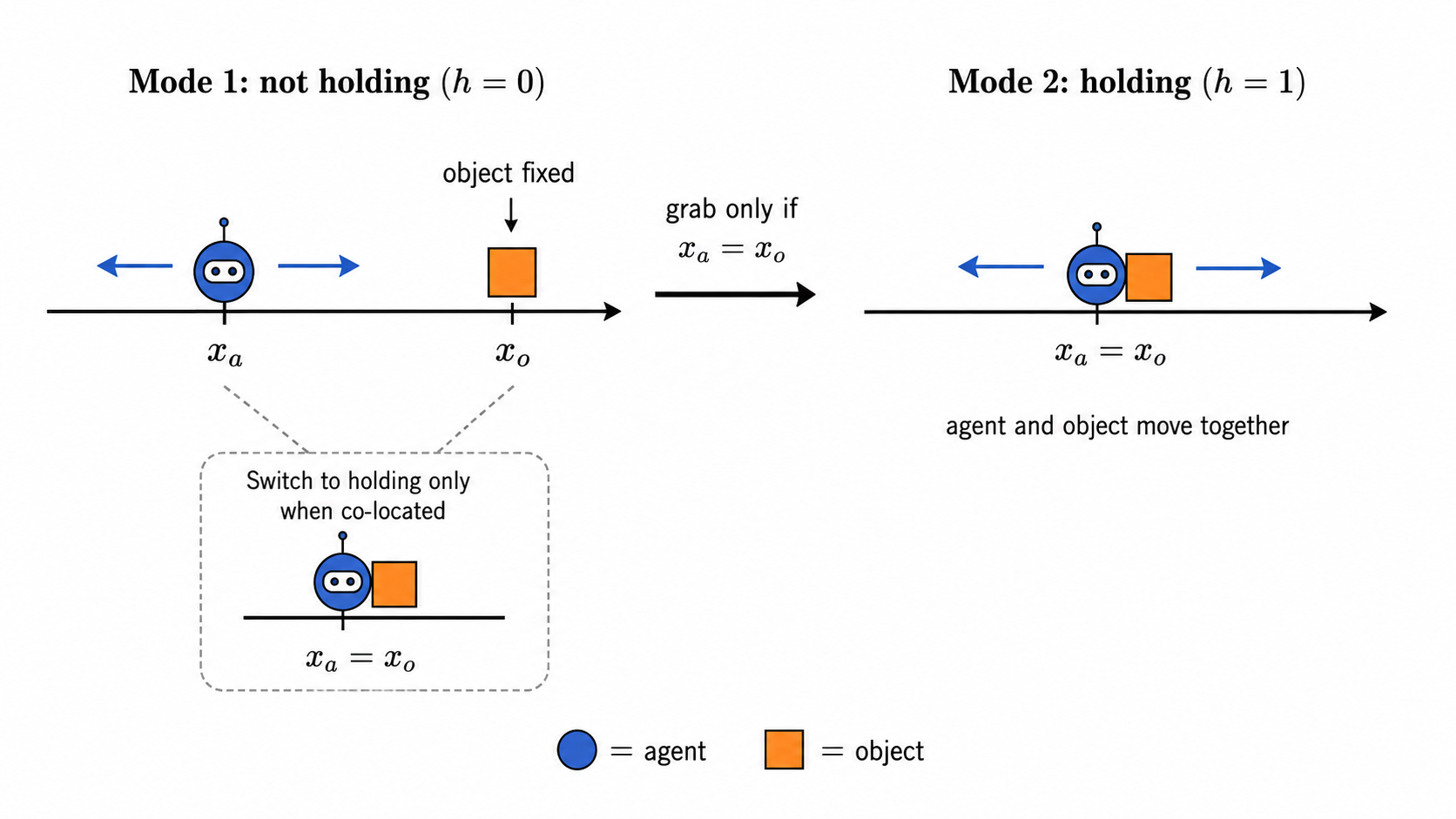}
    \caption{1-D toy example of a contact-rich manipulation task with mode-dependent dynamics.}
    \label{fig:toy_illustrative_example}
\end{figure}

The state is $s=(x_a,x_o,h)$, where $x_a$ denotes the agent position, $x_o$ denotes the object position, and $h\in\{0,1\}$ indicates whether the agent is holding the object. When $h=0$, the agent can move left or right while the object remains fixed. The agent can switch to the holding mode only when it is co-located with the object, i.e., when $x_a=x_o$. Once $h=1$, the object is attached to the agent and moves together with it. We refer to Fig.~\ref{fig:toy_illustrative_example} for a graphical illustration of this toy example.

For completeness, we now write the optimal cost-to-go in an explicit decomposed form. We consider goals in which the object must be held at the target location, so that $h_g=1$ and $x_a^g=x_o^g$. Under unit step cost, if the agent is already holding the object, it only needs to transport it to the target. If the agent is not holding the object, it must first reach the object, switch contact mode, and then transport the object to the target:
\begin{align*}
    d^*(x_a,x_o,h;x_o^g,h_g=1)
    =
    \begin{cases}
        \underbrace{|x_a-x_o^g|}_{\text{transport object to target}},
        & h=1, \\[6pt]
        \underbrace{|x_a-x_o|}_{\text{reach object}}
        +
        \underbrace{1}_{\text{grasp/switch mode}}
        +
        \underbrace{|x_o-x_o^g|}_{\text{transport object to target}},
        & h=0.
    \end{cases}
\end{align*}

Given this ground-truth cost-to-go, we fix $x_o^g=0$ and train a neural approximation $d_{\bm\theta}$ with three losses. The first is an unregularized regression baseline,
\begin{align}
    \mathcal{L}_{\mathrm{base}}(d_{\bm\theta},d^*)
    =
    \mathbb{E}_{s}\left[
    \left(d_{\bm\theta}(s,g)-d^*(s,g)\right)^2
    \right].
    \label{eq_app:toy_baseline}
\end{align}
The second augments this objective with the contact-aware HJB residual in \eqref{eq:ca_hjb},
\begin{align}
    \mathcal{L}_{\mathrm{HJB}}(d_{\bm\theta},d^*)
    =
    \mathcal{L}_{\mathrm{base}}(d_{\bm\theta},d^*)
    +
    \mathcal{L}_{\mathrm{CA\text{-}HJB}}(\bm\theta).
    \label{eq_app:toy_hjb}
\end{align}
The third augments the same regression objective with the full-state Eikonal residual in \eqref{eq:Eik_reg},
\begin{align}
    \mathcal{L}_{\mathrm{Eik}}(d_{\bm\theta},d^*)
    =
    \mathcal{L}_{\mathrm{base}}(d_{\bm\theta},d^*)
    +
    \mathcal{L}_{\mathrm{Eik}}(\bm\theta).
    \label{eq_app:toy_eik}
\end{align}

\begin{figure}[h!]
    \centering
    \includegraphics[width=\linewidth]{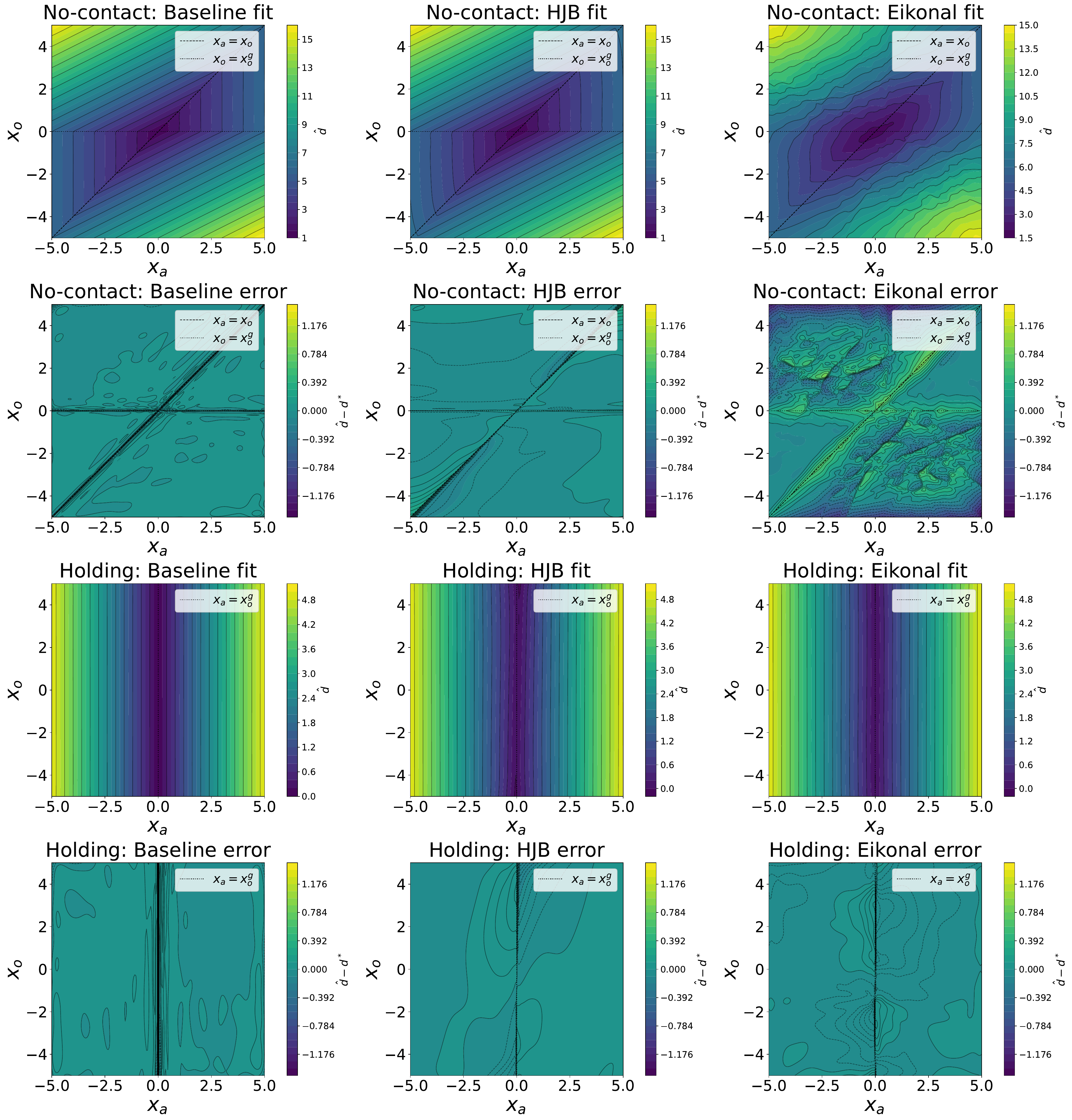}
    \caption{
    Extended visualization of the hybrid toy example in Fig.~\ref{fig:toy_illustrative_example}. Rows show the fitted cost-to-go $\hat d$ and the error $\hat d-d^*$ for the no-contact and holding modes. Columns compare the unregularized baseline in Eq.~\eqref{eq_app:toy_baseline}, the contact-aware HJB regularizer in Eq.~\eqref{eq_app:toy_hjb}, and the full-state Eikonal regularizer in Eq.~\eqref{eq_app:toy_eik}. In the no-contact mode, the Eikonal penalty introduces larger structured errors because the object coordinate affects the long-horizon cost-to-go while being locally uncontrollable. In contrast, the baseline and HJB-regularized solutions more closely recover the piecewise-linear structure of $d^*$. In the holding mode, where the object moves with the agent and the relevant coordinates are locally controllable, all methods recover similar value landscapes. Across methods, residual errors concentrate near the nondifferentiable switching and goal-alignment boundaries, where gradient-based regularization is naturally harder to satisfy.}
    \label{fig:toy_example_extended}
\end{figure}

Fig.~\ref{fig:toy_example_extended} provides a more detailed view of the behavior summarized in the main text. In the no-contact mode, the ground-truth cost-to-go depends on both $x_a$ and $x_o$, but only $x_a$ is locally controllable. The baseline and HJB-regularized models recover the main piecewise-linear structure of $d^*$, with errors mostly localized near nondifferentiable boundaries such as $x_a=x_o$ and $x_o=x_o^g$. This is expected, since the value function has kinks at mode-switching and goal-alignment sets, where a smooth neural approximation cannot exactly match the true nonsmooth geometry.

The full-state Eikonal regularizer behaves differently. In the no-contact mode, it produces a visibly distorted value landscape and structured errors over broad regions of the state space, not only near the nondifferentiable boundaries. This supports Proposition~\ref{prop:eikonal_uncontrollable}, which states that the issue is not merely that nonsmoothness of the value function, but that the Eikonal penalty constrains gradients along the object coordinate even though the object is locally uncontrollable. In the holding mode, where the object moves with the agent, this mismatch disappears and all three losses produce similar fitted landscapes with comparatively small errors. Thus, this extended visualization confirms that the failure of full-state Eikonal regularization is mode-dependent and specifically tied to local uncontrollability.

\subsection{State representation ablation curves}

\begin{figure}[h]
    \centering
    \includegraphics[width=\linewidth]{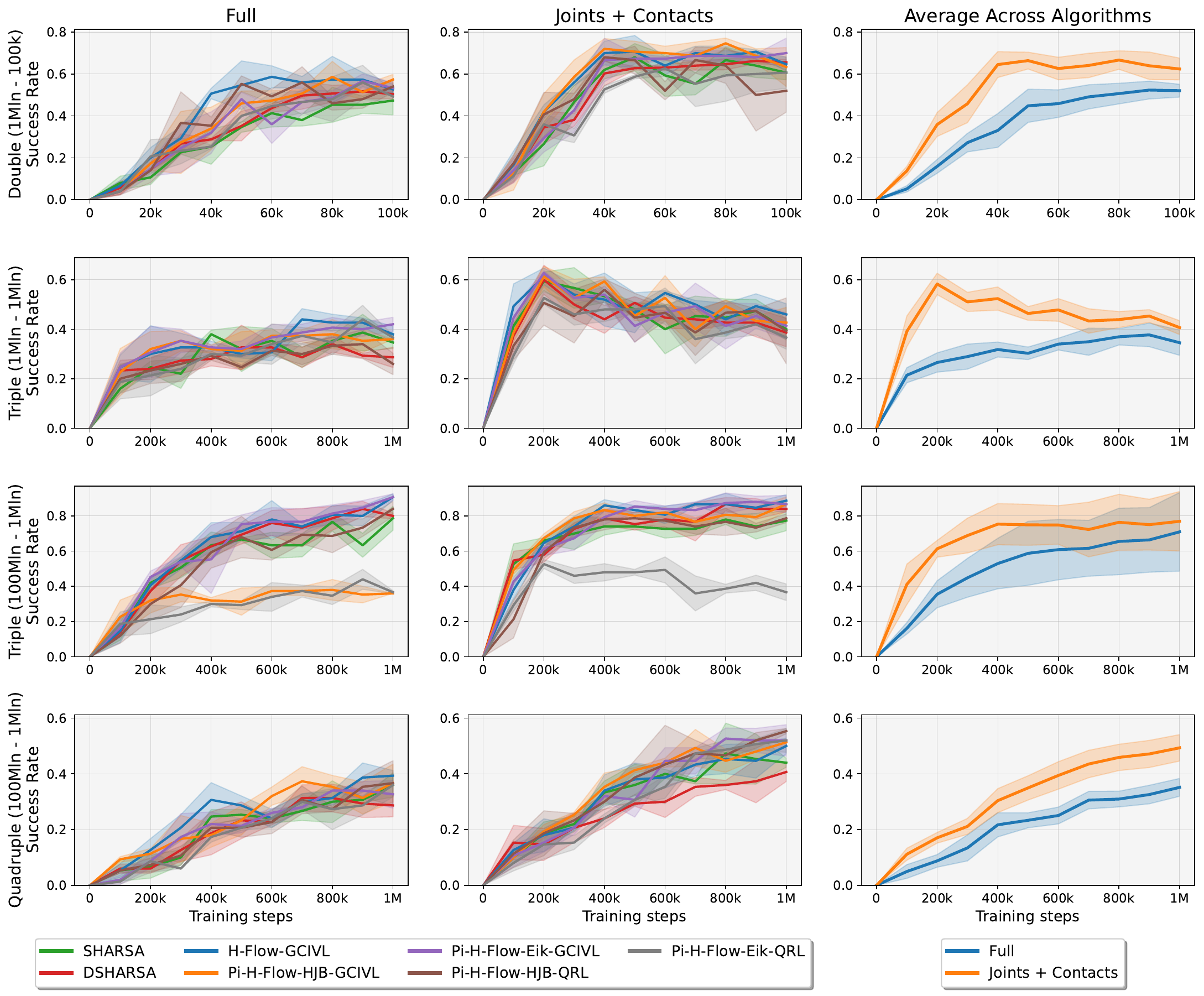}
    \caption{State representation ablation. We compare hierarchical algorithms using the full manipulation state at the low level against variants using only the controllable manipulator-related state. The left column reports results with the full low-level state, the middle column reports results with the proposed ``Joints+Contacts'' representation, and the right column averages performance across algorithms for each representation. The notation in each row title, e.g., \texttt{100Mln-1Mln}, denotes the dataset size and number of training steps, respectively. Evaluation is performed as in Fig.~\ref{fig:ogbench_flat}. Curves report the mean and standard deviation across seeds.}
    \label{fig:app_ogbench_hierarchical_learning_curves}
\end{figure}

Fig.~\ref{fig:app_ogbench_hierarchical_learning_curves} provides the full training curves corresponding to the ablation summarized in Fig.~\ref{fig:ogbench_hierarchical}. The left and middle columns report the performance of each hierarchical method when the low-level MDP is defined on the full manipulation state and on the proposed controllable ``Joints+Contacts'' representation, respectively. The latter excludes object-state coordinates from the low-level problem. The right column averages performance across algorithms to isolate the effect of the low-level state representation.

Across environments, the controllable representation generally leads to faster learning and higher final success rates. This improvement is visible across all the environments, where using the full manipulation state often slows learning or plateaus at lower success rates. These trends support the claim that removing object-pose coordinates from the low-level problem simplifies goal-reaching and better aligns the low-level learner with locally controllable dynamics.

\section{Real-world Experiments}
\label{sec_app:real_world_experiments}

In this section, we provide additional details on the real-world experiments presented in the main text in Section~\ref{sec:experiments}, Fig.~\ref{fig:setup_real_world_experiments}. We first describe the experimental setup and data-collection procedure, and then present qualitative results that complement the quantitative evaluation reported in the main paper.

\subsection{Setup and data collection}

\begin{figure}[h]
    \centering
    \includegraphics[height=2cm, width=\linewidth]{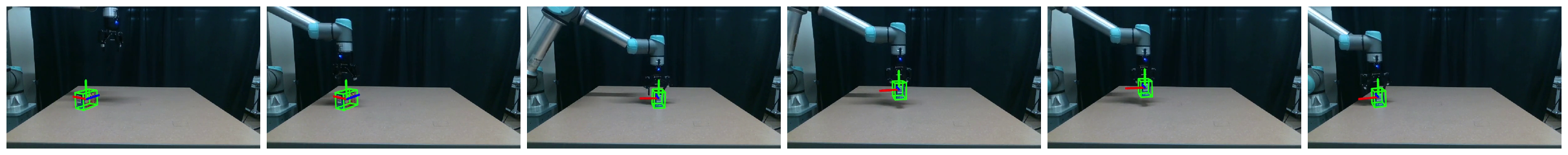}
    \caption{Data collection on the real-world pick-and-place setup. The task is performed with a UR5e robot, while RGB-D observations are recorded using an Intel RealSense D415 camera. The pose of the manipulated object is estimated using FoundationPose~\citep{wen2024foundationpose}.}
    \label{fig_app:data_collection}
\end{figure}

We consider a tabletop pick-and-place task in which a UR5e robot must grasp an object and move it between randomly sampled configurations and a target goal region. Object pose estimation is performed from RGB-D observations captured by an Intel RealSense D415 camera using FoundationPose~\citep{wen2024foundationpose}. The policy input consists of robot proprioception, including joint positions and velocities, together with the end-effector yaw, gripper state, and estimated object pose.

To collect offline interaction data, we design a scripted behavior policy based on an MPC controller~\citep{manganaris2026graph}. The controller uses the estimated state of the robot-object system to generate pick-and-place trajectories between random object configurations and the goal state. An example rollout of this data-collection policy is shown in Fig.~\ref{fig_app:data_collection}. We collect approximately $80$k transition samples and subsequently filter the dataset during post-processing, removing failed executions and poor-quality demonstrations. This yields a final offline dataset of approximately $60$k samples used for training.

\subsection{Extended summary of the results}
\vspace{-0.3cm}
\begin{table}[h]
    \centering
    \caption{Real-world pick-and-place success rates over $20$ evaluation episodes. Policies are trained on approximately $60$k interaction steps for $300$k gradient steps.}
    \label{tab:real_world_results}
    \begin{tabular}{lccc}
        \toprule
        Method & $3$cm & $5$cm & $10$cm \\
        \midrule
        GCIVL & $25$ & $40$ & $45$ \\
        HJB-GCIVL & $40$ & $50$ & $65$ \\
        H-Flow-GCIVL & $25$ & $40$ & $40$ \\
        Pi-H-Flow-HJB-GCIVL & $\mathbf{50}$ & $\mathbf{60}$ & $\mathbf{75}$ \\
        \bottomrule
    \end{tabular}
    \vspace{0.3em}
    
    \small{Success rates are reported in percent for object-position thresholds of $3$cm, $5$cm, and $10$cm.}
\end{table}

Table~\ref{tab:real_world_results} provides a compact summary of the real-world evaluation results discussed in Section~\ref{sec:experiments} of the main text. Policies are trained on approximately $60$k interaction steps for $300$k gradient steps and evaluated over $20$ episodes on the physical system. Success rates are reported under three object-position thresholds, corresponding to $3$cm, $5$cm, and $10$cm Euclidean distance from the center of the table. Overall, the results show that the HJB-style regularizer improves performance over the GCIVL baseline, and that performance improves further when physics-informed value learning is combined with the proposed hierarchical decomposition. In particular, Pi-H-Flow-HJB-GCIVL achieves the strongest performance across all evaluation thresholds.

\begin{figure}[h]
    \centering

    \begin{subfigure}[t]{\linewidth}
        \centering
        \includegraphics[height=2cm, width=\linewidth]{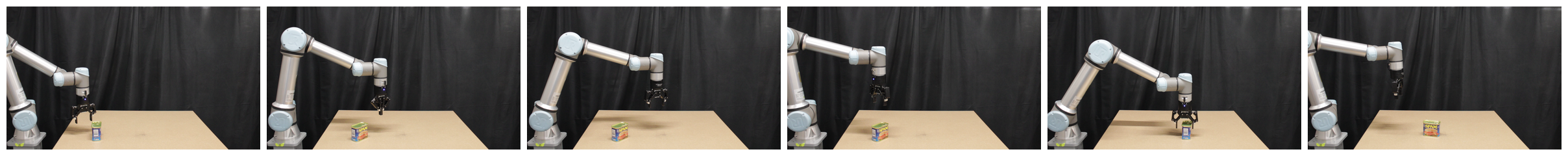}
        \caption{Recovery after an initially misaligned grasp attempt.}
        \label{fig:recovery_after_fail}
    \end{subfigure}

    \vspace{0.4em}

    \begin{subfigure}[t]{\linewidth}
        \centering
        \includegraphics[height=2cm, width=\linewidth]{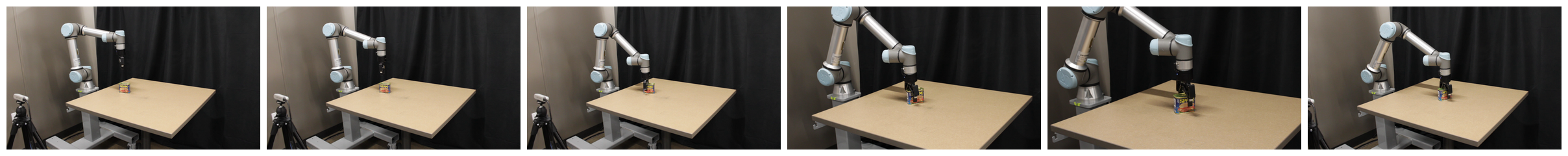}
        \caption{Successful object transport through sliding.}
        \label{fig:sliding_mode}
    \end{subfigure}

    \vspace{0.4em}

    \begin{subfigure}[t]{\linewidth}
        \centering
        \includegraphics[height=2cm, width=\linewidth]{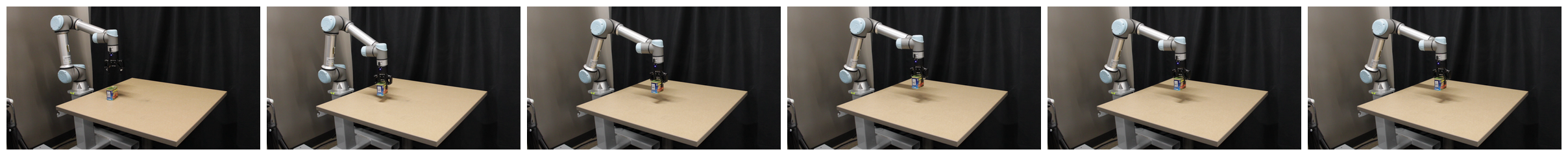}
        \caption{Failure case where the policy stalls before placement.}
        \label{fig:stuck_before_failing}
    \end{subfigure}

    \vspace{0.4em}

    \begin{subfigure}[t]{\linewidth}
        \centering
        \includegraphics[height=2cm, width=\linewidth]{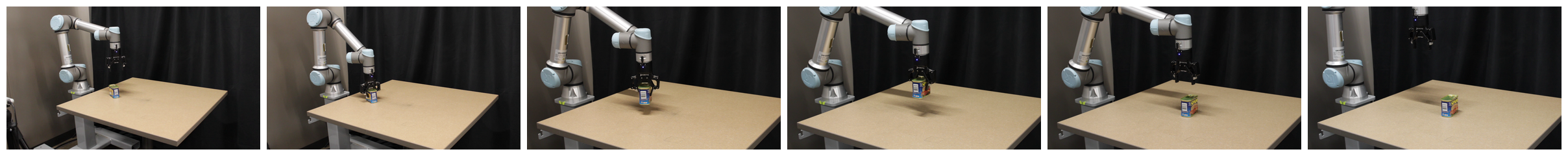}
        \caption{Smooth successful pick-and-place rollout.}
        \label{fig:success}
    \end{subfigure}

    \vspace{0.4em}

    \begin{subfigure}[t]{\linewidth}
        \centering
        \includegraphics[height=2cm, width=\linewidth]{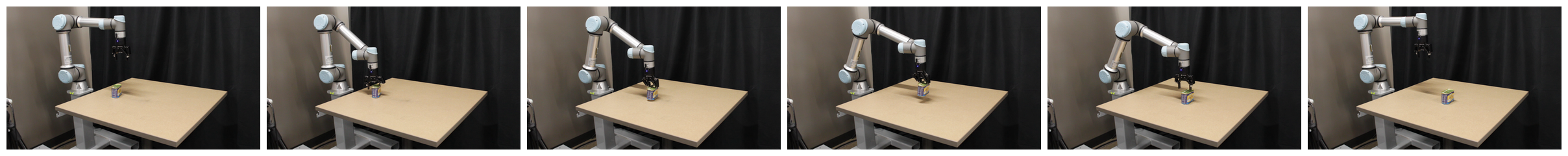}
        \caption{Successful recovery after grasping near the object rim.}
        \label{fig:success_edge_pick}
    \end{subfigure}

    \caption{Qualitative examples of real-world rollouts collected from the learned policy. Each row shows temporally ordered frames from a representative trajectory, illustrating successful behaviors, recovery from imperfect grasps, alternative interaction modes, and representative failure cases.}
    \label{fig:real_world_qualitative_rollouts}
\end{figure}

We further provide qualitative examples of the learned behavior in Fig.~\ref{fig:real_world_qualitative_rollouts}. These rollouts are collected from an HJB-GCIVL policy and are intended to complement the quantitative results by illustrating both successful behaviors and representative failure modes. Additional qualitative results will be included in the supplementary video.

The rollout in Fig.~\ref{fig:recovery_after_fail} shows a recovery behavior: the agent initially misses the object due to poor end-effector alignment, but subsequently repositions, grasps the object, and completes the task. 

Fig.~\ref{fig:sliding_mode} shows a successful rollout in which the agent moves the object by sliding it along the table rather than directly lifting and placing it. This behavior was not explicitly present in the demonstrated dataset, suggesting that the learned policy can exploit alternative interaction modes. 

Fig.~\ref{fig:stuck_before_failing} shows a representative failure case in which the agent gets stuck in mid-air before placing the object. In this configuration, we observed that the policy sometimes recovers, while in other cases external intervention is required. 

Finally, Fig.~\ref{fig:success} and Fig.~\ref{fig:success_edge_pick} show two successful rollouts. The first corresponds to a smooth pick-and-place behavior similar to the demonstrated trajectories, whereas the second shows a less typical strategy in which the robot grasps the object near its rim due to imperfect object--end-effector alignment, but still recovers and completes the task.

These preliminary real-world results provide evidence that the proposed approach can produce robust and adaptive manipulation behaviors on a physical system, while also highlighting remaining failure modes that motivate future work on safety, compliance, and more structured contact-aware control.

\end{document}